\RequirePackage{fix-cm}

\documentclass[smallextended]{svjour3}      

\smartqed

\usepackage{caption}
\usepackage{subcaption}
\usepackage{framed} 
\usepackage[flushleft]{threeparttable}
\usepackage{graphicx}
\usepackage{amsfonts}
\usepackage{amsmath}
\usepackage{amssymb}
\usepackage{mathtools}
\usepackage{float}
\usepackage[]{geometry}
\usepackage{color}
\usepackage[dvipsnames]{xcolor}
\usepackage[toc,page]{appendix}
\usepackage{enumitem,xcolor}

\begin{document}

\title{A review on distance based  time series classification}

\author{Amaia Abanda   \and Usue Mori  \and Jose A. Lozano}

\institute{Amaia Abanda$^{1,2}$  \at
              $^{1}$ Basque Center for Applied Mathematics (BCAM) \\
              Mazarredo Zumarkalea, 14, 48009 Bilbo, Spain \\
               $^{2}$ Intelligent Systems Group (ISG)\\
              Department of Computer Science and Artificial Intelligence, University of the Basque Country UPV/EHU\\
              Manuel de Lardizabal 1, 20018 Donostia-San Sebastian, Spain\\
              \email{aabanda@bcamath.org}           
           \and
            Jose A. Lozano$^{1,2}$ \at
              $^{1}$ Basque Center for Applied Mathematics (BCAM) \\
              Mazarredo Zumarkalea, 14, 48009 Bilbo, Spain \\
              $^{2}$ Intelligent Systems Group (ISG)\\
              Department of Computer Science and Artificial Intelligence, University of the Basque Country UPV/EHU\\
              Manuel de Lardizabal 1, 20018 Donostia-San Sebastian, Spain\\
              \email{ja.lozano@ehu.eus}                          
 		 \and  
 		 Usue Mori$^{2,3}$ \at
 		  $^{2}$ Intelligent Systems Group (ISG)\\
              Department of Computer Science and Artificial Intelligence, University of the Basque Country UPV/EHU\\
              Manuel de Lardizabal 1, 20018 Donostia-San Sebastian, Spain\\ 		
               $^{3}$ Department of Applied Mathematics, Statistics and Operational Research\\
               University of the Basque Country UPV/EHU \\
               Barrio Sarriena, s/n, 48940 Leoia, Spain \\
              \email{usue.mori@ehu.es}    
}        %

\date{Received: date / Accepted: date}

\maketitle

\begin{abstract}
\small{
Time series classification is an increasing research topic due to the vast amount of time series data that are being created over a wide variety of fields. The particularity of the data makes it a challenging task and different approaches have been taken, including the distance based approach. 1-NN has been a widely used method within distance based time series classification due to it simplicity but still good performance. However, its supremacy may be attributed to being able to use specific distances for time series within the classification process and not to the classifier itself. With the aim of exploiting these distances within more complex classifiers, new approaches have arisen in the past few years that are competitive or which outperform the 1-NN based approaches.  In some cases, these new methods use the distance measure to transform the series into feature vectors, bridging the gap between time series and traditional classifiers. In other cases, the distances are employed to obtain a time series kernel and enable the use of kernel methods for time series classification. One of the main challenges is that a kernel function must be positive semi-definite, a matter that is also addressed within this review. The presented review includes a taxonomy of all those methods that aim to classify time series using a distance based approach, as well as a discussion of the strengths and weaknesses of each method. }

\end{abstract}

\newpage

\section{Introduction}
\label{intro}

Time series data are being generated everyday in a wide range of application domains, such as bioinformatics, financial fields, engineering, etc  \cite{Keogh2002}. They represent a particular type of data due to their \textit{temporal nature}; a time series is, in fact, an ordered sequence of observations of finite length which are usually taken through time, but may also be ordered with respect to another aspect, such as space. With the growing amount of recorded data, the interest in researching this particular data type has also increased, giving rise to a vast amount of new methods for representing, indexing, clustering and classifying time series, among other tasks \cite{Esling2012}. This work focuses on time series classification (TSC), and in contrast to traditional classification problems, where the order of the attributes of the input objects is irrelevant, the challenge of TSC consists of dealing with  temporally correlated attributes, i.e., with inputs in which each $x_{i}$ is defined by a complete ordered sequence \cite{Bagnall2017}\cite{Fu2011}.

Time series classification methods can be divided into three main categories \cite{Xing2010}: feature based, model based and distance based methods. In feature based classification methods, the time series are transformed into  feature vectors and then classified by a conventional classifier such as a neural network or a decision tree. Some methods for feature extraction include spectral methods such as discrete Fourier transform (DFT) \cite{ChristosFaloutsosM.Ranganathan1994} or discrete wavelet transform (DWT), \cite{Popivanov2002} where features of the frequency domain are considered, or singular value decomposition (SVD) \cite{FlipKornH.V.Jagaciish1997}, where eigenvalue analysis is carried out in order to find an optimal set of features. On the other hand, model based classification assumes that all the time series in a class are generated by the same underlying model, and thus a new series is assigned with the class of the model that  best fits it. Some model based approaches are formed on using auto-regressive models  \cite{Bagnall2014}\cite{Corduas2008} or hidden Markov models \cite{Smyth1997}, among others.  Finally, distance based methods are those in which a (dis)similarity measure between series is defined, and then these distances are introduced in some manner within  distance-based classification methods such as the k-nearest neighbour classifier (k-NN) or Support Vector Machines (SVMs). This work focuses on this last category, distance based classification of time series. 

\begin{figure}[h]
\centering
\includegraphics[scale=0.5]{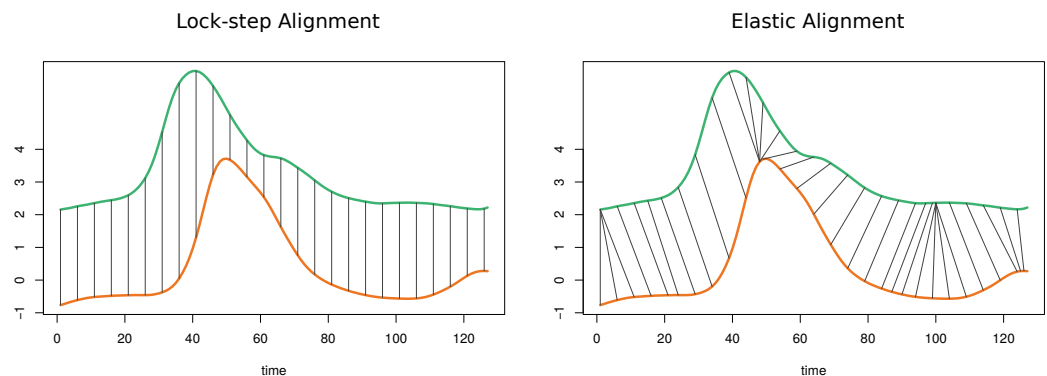}
\caption{\small{Mapping of Euclidean distance (lock-step measure) vs. mapping of DTW distance (elastic measure)}}
\vspace{0.4cm}

\label{fig:align}
\end{figure}

Until now, almost all the research in distance based classification has been oriented to defining different types of distance measures and then exploiting them within k-NN classifiers. Due to the temporal (ordered) nature of the series, the high dimensionality, the noise and the possible different lengths of the series in the database, the definition of a proper distance measure is a key issue in distance based time series classification. One of the ways to categorize time series distance measures in shown in Figure \ref{fig:align};  \textit{Lock-step measures} refer to those distances that compare the $i$th point of one series to the $i$th point of another (e.g., Euclidean distance), while \textit{elastic measures} aim to create a non-linear mapping in order to align the series and allow comparison of one-to-many points (e.g., Dynamic Time Warping \cite{Berndt1994}). These two types of measures consider that the important aspect to define the distance is the shape of the series, but there are also structure based or edit based measures, among others  \cite{Esling2012}. In this sense, different distance measures are able to capture different types of dissimilarities, and there is no unique best distance for all cases. Nevertheless, the experimentation in 	\cite{Esling2012}\cite{Xing2010}\cite{Wang2013}\cite{Chen2013}\cite{Ding2008}\cite{Lines2015}\cite{Xi2006} has shown that, on average, the DTW distance seems to be particularly difficult to beat.

One of the simplest ways to exploit a distance measure within a classification process is by employing k-NN classifiers. One could expect that a more complex classifier would outperform the performance of the 1-NN, so the bad performance of these complex classifiers may be attributed to the inability of the classifiers to deal with the temporal nature of the series using the default settings. On the other hand, it is  known that the underlying distance is crucial to the performance of the 1-NN classifier \cite{Tan2005} and, hence, the high accuracy of 1-NN classifiers may arise from the efficiency of the time series distance measures -which take into consideration the temporal nature- for classification.   In this way, methods that exploit the potential of these distances within more complex classifiers have emerged in the past few years \cite{Kate2015}\cite{Jalalian2013}\cite{Marteau2014},  achieving performances  that are competitive or outperform the classic 1-NN.

These new approaches aim to take advantage of the existing time series distances to exploit them within more complex classifiers. We have differentiated between two new ways of using distance measures in the literature: the first employs the distance to obtain a new feature representation of the series \cite{Kate2015}\cite{Iwana2017}\cite{Hills2014}, i.e., a representation of the series as an order-free vector, while the second uses the distance to obtain a kernel \cite{Gudmundsson2008}\cite{Cuturi2007}\cite{Marteau2014}, i.e., a similarity between the series that will then be used within a kernel method. Both approaches have achieved competitive classification results and, thus, different variants have arisen \cite{Jeong2015}\cite{Zhang2010a}\cite{Lods}. The purpose of this review is to present a taxonomy of all those methods which are based on  time series distances for classification. At the same time, the strengths and shortcomings of each approximation are discussed in order to give a general overview of the current research directions in distance based time series classification.

The rest of the paper is organized as follows: in Section \ref{sec:2} the taxonomy of the reviewed methods is presented, as well as a brief description of the methods in each category. In Section \ref{sec:6} a discussion on the approaches in the taxonomy is presented, where we draw our conclusions and specify some future directions.

\vspace{0.5cm}
\section{A taxonomy of distance based time series classification}
\label{sec:2}

\begin{figure}[h]
\begin{center}
\includegraphics[scale=0.8]{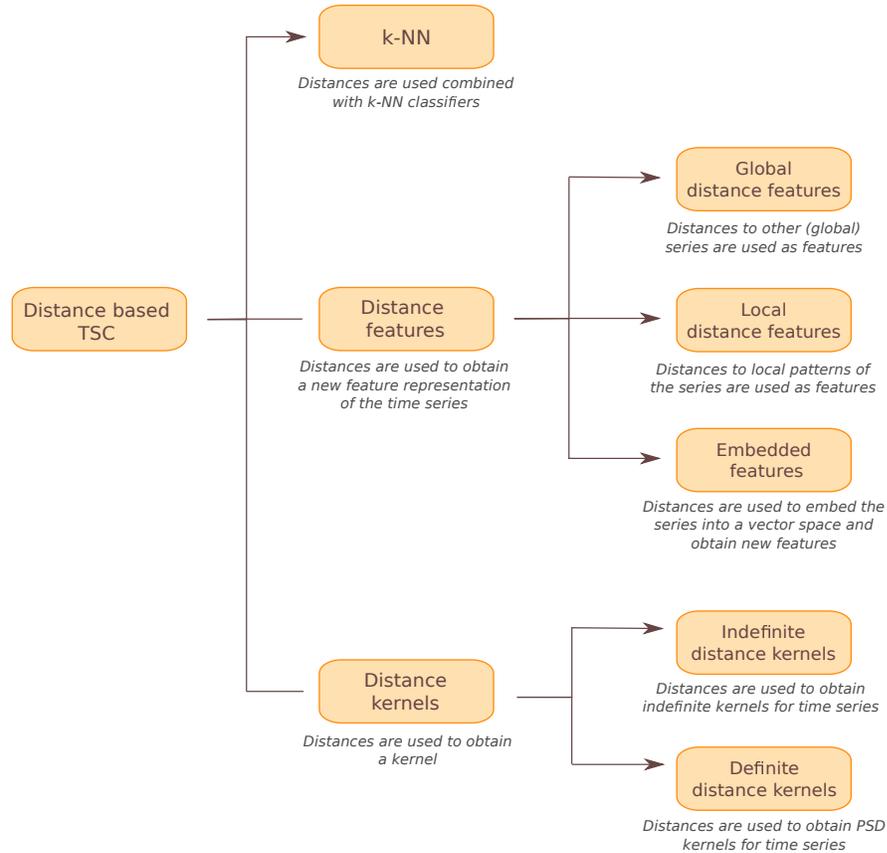}
\caption{Taxonomy of distance based time series classification}
\label{fig:tax}
\end{center}
\end{figure}

As mentioned previously, the taxonomy we propose intends to include and categorize all the distance based approaches for time series classification. A visual representation of the taxonomy can be seen in Figure~\ref{fig:tax}. From the most general point of view, the methods can be divided into three main categories: in the first one, the distances are used directly in conjunction with k-NN classifiers; in the second one, the distances are used to obtain a new representation of the series by transforming them into features vectors, while in the third one, the distances are used to obtain kernels for time series.

\vspace{0.3cm}

\vspace{0.5cm}

\subsection{k-Nearest Neighbour}
\label{sec:3}

This approach employs the existing time series distances within k-NN classifiers. In particular, the 1-NN classifier has mostly been used in time series classification due to its simplicity and competitive performance \cite{Ding2008}\cite{Lines2012}. Given a distance measure and a time series, the 1-NN classifier predicts the class of this series as the class of the object closest to it from the training set. Despite the simplicity of this rule, a strength of the 1-NN is that, as the size of the training set increases, the 1-NN classifier guarantees an error lower than two times the Bayes error \cite{Cover1967}. Nevertheless, it is worth mentioning that it is very sensitive to noise in the training set, which is a common characteristic of time series datasets. This approach has been widely applied in time series classification, as it achieves, in conjunction with the DTW distance, the best accuracy ever reached in many benchmark datasets. As such, quite a few studies and reviews include the 1-NN in the time series literature \cite{Bagnall2017}\cite{Wang2013}\cite{Lines2015}\cite{Kaya2015}, and, hence, it is not going to be   further detailed in this review.

\vspace{0.4cm}
\subsection{Distance features}
\label{sec:4}
\vspace{0.1cm}

In this group, we include the methods that employ a time series distance measure to obtain a new representation of the series in the form of feature vectors. In this manner, the series are transformed into feature vectors (order-free vectors in $\mathbb{R}^{N}$), overcoming many specific requirements that are encountered in time series classification, such as dealing with ordered sequences or handling instances of different lengths. The main advantage of this approach is that it bridges the gap between time series classification and conventional classification, enabling the use of general classification  algorithms designed for vectors, while taking advantage of the potential of the time series distances. In this manner, calculating the distance features can be seen as a preprocessing step and, thus, the transformation can be used in combination with any classifier. Note that even if these methods also obtain  some features from the series, they are not considered within feature based time series classification, but within distance based time series classification. The reason is that the methods in feature based time series classification obtain features that contain information about the series themselves, while distance features contain information relative to their relation with the other series. Three main approaches are distinguished within this category: those that directly employ the vector made up of the distances to other series as a feature vector, those that define the features using the distances to some local patterns, and those that use the distances after embedding the series into some vector space.

\vspace{0.3cm}
\subsubsection{Global distance features}
\label{sec:4.1}

\begin{figure}[h]
\begin{center}
\vspace{0.2cm}
\includegraphics[scale=0.65]{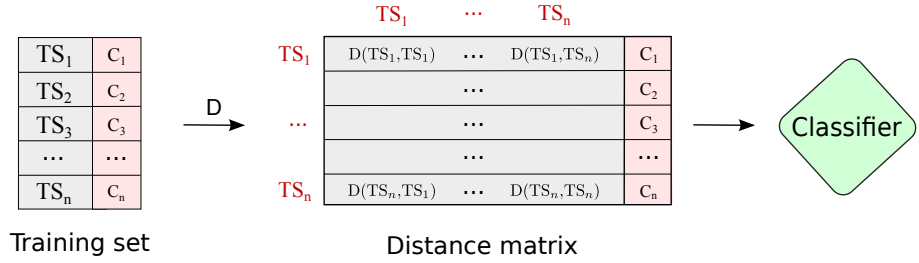}
\caption{A visual representation of the global distance features method.}
\label{fig:asfea}
\end{center}
\vspace{0.3cm}
\end{figure}

The main idea behind the methods in this category is to convert the time series into feature vectors by employing the vector of distances to other series as the new representation. Firstly, the distance matrix is built by calculating the distances between each pair of samples, as shown in Figure \ref{fig:asfea}. Then, each row of the distance matrix is used as a feature vector describing a time series, i.e., as input for the classifier. It is worth mentioning that this approach is a general approach (not specific for time series) but becomes specific when a time series distance measure is used.  Learning with the distance features is also known as learning in the so-called  \textit{dissimilarity space} \cite{Pekalska}. For more details on learning with global distance features in a general context, see \cite{Pekalska}\cite{Chen2009}\cite{Paclik2001}\cite{Graepel1999}. 

Even if learning with distance features is a general solution, it is particularly advantageous for time series; the distance to each series  is understood as an independent dimension and the series can be seen as vectors in an Euclidean space. The potential of this approach is that this new representation enables the use of conventional classifiers that are designed for feature vectors, while it takes advantage of the existing time series distances.  However, learning from the distance matrix has some important drawbacks; first, the distance matrix must be calculated, which may be costly depending on the complexity of the distance measure. Then, once the distance matrix has been calculated, learning a classifier with it may also incur large computational cost, due to the possible large size of the training set.  Note that, in the prediction stage, the consistent treatment of a new time series is straightforward -just the distances from the new series to the series in the training set have to be computed- but it can also become computationally expensive depending on the distance measure. Henceforth, given a distance measure $d$, we will refer to the methods employing the corresponding distance features as  DF$_{d}$  .

After a brief introduction of the distance based features, a summary of the methods employing them is now presented. Gudmundsson \textit{et al.} \cite{Gudmundsson2008} made the first attempt at investigating the feasibility of using a time series distance measure within a more complex classifier than the k-NN. In particular, they aimed at taking advantage of the potential of Support Vector Machines (SVMs) on the one hand, and of Dynamic Time Warping (DTW) on the other. First, they converted the DTW distance measure into two DTW-based similarity measures, shown in equation (\ref{gdtw}). Then, they employed the distance features obtained from these similarity measures,  DF$_{GDTW}$ and DF$_{NDTW}$, in combination with SVMs for classification.

\vspace{0.4cm}
\small
\begin{equation}
\label{gdtw}
 GDTW(TS_{i}, TS_{j}) = \exp \left(  - \frac{DTW( TS_{i}, TS_{j} )^2}{ \sigma^2} \right), \hspace*{0.5cm} NDTW(TS_{i}, TS_{j}) = -DTW( TS_{i}, TS_{j})
 \vspace{0.5cm}
 \end{equation}\normalsize
where $\sigma > 0 $ is a free parameter and $TS_{i},TS_{j}$ are two time series. They concluded that the new representation in conjunction with SVMs is competitive with the benchmark 1-NN with DTW. 

In \cite{Jalalian2013}, Jalalian \textit{et al.} introduced a \textit{Two-step DTW-SVM} classifier where the DF$_{DTW}$ are used in order to solve a multi-class classification problem. In the prediction stage, the new time series is represented by the distance to all the series in the training set and a voting scheme is employed to classify the series using all the trained SVMs in a one-vs-all schema. They concluded that even if DF$_{DTW}$ achieves acceptable accuracy values, the prediction of new time series is too slow for real world applications when the training set is relatively big.

Additionally, based on the potential of using distances as features for time series classification, Kate \textit{et al.} \cite{Kate2015} carried out a comprehensive experimentation in which different distance measures are used as features within SVMs. In particular, they tested not only DF$_{DTW}$  but also a constrained version DF$_{DTW-R}$ (a window-size constrained version of DTW which is computationally faster \cite{Sakoe1978}), features obtained from the Euclidean distance DF$_{ED}$ and also concatenations of these distance features with other feature based representations. In their experimentation, they showed that even the DF$_{ED}$, when used as features with SVMs, outperforms the accuracy of 1-NN classifier based on the same Euclidean distance.  An extension of \cite{Kate2015} was presented in \cite{Giusti2017} by Giusti \textit{et al.}, who argued that not all relevant features can be described in the time domain (frequency domain can be more discriminative, for example)  and added new representations to the set of features. Specifically, they generalized the concept of distance features to other domains and employed 4 different representations of the series  with 6 different distance measures, giving rise to 24 distance features. That is, for each representation of the series $R_{i}$, $i=1,\dots, 4$, they computed 6 different distance features DF$_{d_{1}}^{R_{i}}, \dots, \text{DF}_{d_{6}}^{R_{i}}$. In their experimentation on 85 datasets from the UCR\footnotemark  \hspace{0.03cm} they showed that using representation diversity improves the classification accuracy. Finally, in their work about early classification of time series, Mori \textit{et al.} \cite{Mori2017} benefit from Euclidean distance features DF$_{ED}$ in order to classify the series with SVMs and Gaussian Processes \cite{Rasmussen2006}.

\footnotetext{UCR is a repository of time series datasets  \cite{Chen2015a} which is often used as a benchmark for evaluating time series classification methods. These datasets are greatly varied with respect to their application domains, time series lengths, number of classes, and sizes of the training and testing sets.}

With the aim of addressing the limitation of the  high computational cost of the DTW distance,  Janyalikit \textit{et al.} \cite{Goebel2016} proposed the use of a fast lower bound for the DTW algorithm, called LB\_Keogh \cite{Keogh2005}. Employing DF$_{LB\_Keogh}$ with SVMs, Janyalikit \textit{et al.} showed in their experimentation on 47 UCR datasets that their method speeds  the classification task up by a large margin, while maintaining the accuracies comparing with the state-of-art DF$_{DTW-R}$ proposed in \cite{Kate2015}.

As previously mentioned, another weakness of using distances as features is the high dimensionality of the distance matrix, since for $n$ instances a $n \times n$  matrix is used as the input to the classifier. In view of this, Jain \textit{et al.} \cite{Goebel2015} proposed  a dimensionality reduction approach using Principal Component Analysis (PCA) in order to keep only those dimensions that retain the most information. In their experimentation they compare the use of DF$_{DTW}$ with the reduced version of the same matrix, DF$_{DTW+PCA}$  in combination with SVMs. They showed that PCA can have a consistent positive effect on the performance of the classifier but this effect seems to be dependent of the choice of the kernel function applied in the SVM. Note that for prediction purposes, they transform the new time series using the PCA projection learned from the training examples and, hence, the prediction process will also be significantly faster.

Another dimensionality reduction approach used in these cases is prototype selection, employed by Iwana \textit{et al.}  \cite{Iwana2017}. The idea is to select a set of $k$ reference time series, called prototypes, and compute only the distances from the series to the $k$ prototypes. The authors pointed out that the distance features let each feature to be treated independently and, consequently, prototype selection can be seen as a feature selection process. As shown in \cite{Goebel2015}, this dimensionality reduction technique not only speeds up the training phase but also the prediction of new time series. The proposed method uses the AdaBoost \cite{Freund1997} algorithm, which is able to select discriminative prototypes and combine a set of weak learners. They experimented with DF$_{DTW+PROTO}$ and analyzed different prototype selection methods. 

To conclude this section, a summary of the reviewed methods of \textit{Global distance features}  for TSC can be found in Table \ref{tab:summary_DF}.

\begin{center}
{\renewcommand{\arraystretch}{1.1} 
\begin{threeparttable}
\captionof{table}{Summary of global distance feature approaches} 
\begin{tabular}[h]{|l |l |l |l |}
\hline
 \textbf{Authors} & \textbf{Features} & \textbf{Classifier} & \textbf{Datasets} \\ \hline
 Gudmundsson \textit{et al.}\cite{Gudmundsson2008}&\small{DF$_{GDTW}$, DF$_{NDTW}$}&SVMs &20 UCR \\ 
 Jalalian \textit{et al.}  \cite{Jalalian2013} & \small{DF$_{DTW}$} & SVMs & 20 UCR \\ 
Kate \textit{et al.}\cite{Kate2015} & \footnotesize{DF$_{ED}-\text{DF}_{DTW}-\text{DF}_{DTW-R}- SAX $} & SVMs & 47 UCR  \\  
Giusti \textit{et al.} \cite{Giusti2017} & \small{DF$_{d_{1, \dots, 6}}^{R_{1, \dots, 4}}$}  & SVMs & 85 UCR  \\   
  Mori  \textit{et al.}   \cite{Mori2017}    & \small{DF$_{ED}$} & GPs, SVMs &  45 UCR \\   
  Janyalikit \textit{et al.} \cite{Goebel2016} & \small{DF$_{LB\_Keogh}$} & SVMs & 47 UCR  \\ 
  Jain \textit{et al.}	\cite{Goebel2015}   & \small{DF$_{DTW+PCA}$} & SVMs & 42 UCR \\ 
  Iwana \textit{et al.} \cite{Iwana2017} & \small{DF$_{DTW+PROTO}$} & Adaboost & 1 (UNIPEN) \\ 
\hline 
\end{tabular}
\label{tab:summary_DF}

\end{threeparttable}
}
\end{center}

\vspace{0.3cm}
\subsubsection{Local distance features}
\label{sec:4.2}

\begin{figure}[h]
\begin{center}
\includegraphics[scale=0.65]{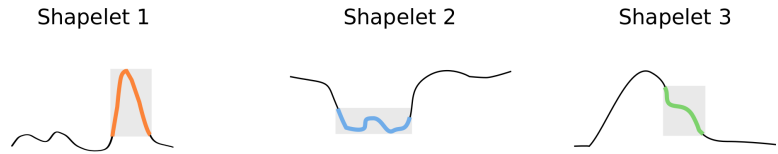}
\caption{Visual representation of three shapelets which belong to three time series belonging to different classes. These shapelets are identified as being representative of class membership.}
\label{fig:shapelets_serie}
\end{center}
\end{figure}

In this section, instead of using distances between entire series, distance to some local patterns of the series are used as features. Instead of assuming that the discriminatory characteristics of the series are global, the methods in this section consider that they are local. As such, the methods in this category employ the so-called \textit{shapelets} \cite{Ye2009}, subsequences of the series that are identified as being representative of the different classes. An example of three shapelets belonging to different time series can be seen in Figure \ref{fig:shapelets_serie}. An important advantage of working with shapelets is their interpretability, since an expert may understand the meaning of the obtained shapelets. Since shapelets are, by definition, subsequences, the methods employing shapelets are not a priori applicable to other types of data.  However, it is worth mentioning that the original shapelet discovery technique, proposed by Ye \textit{et al} \cite{Ye2009}, is carried out by enumerating all possible candidates (all possible subsequences of the series) and using a measure based on information theory that takes $O(n^2m^4)$, where $n$ is the number of time series and $m$ is the length of the longest series. Thereby, most of the work related to shapelets has focused on speeding up the shapelet discovery process \cite{He2012}\cite{Mueen2011}\cite{Rakthanmanon2013}\cite{Ye2011} or on proposing new shapelet learning methods \cite{Grabocka2014a}. However, we will not focus on that but on how shapelets can be used within distance based classification.

\begin{figure}[h]
\begin{center}
\includegraphics[scale=0.65]{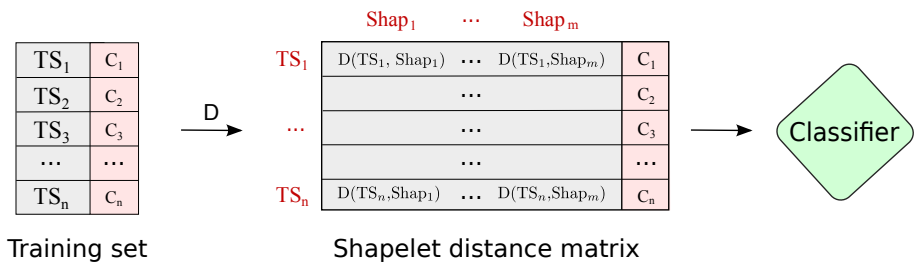}
\caption{Example  of the local distance features methods using ST.}
\label{fig:shapelets_stages}
\end{center}
\end{figure}

\vspace{-0.5cm}
Building on the achievements of shapelets in classification, Lines \textit{et al} \cite{Lines2012} introduced the concept of \textit{Shapelet Transform} (ST). First,  the $k$ most discriminative (over the classes) shapelets are found using one of the methods referenced above. Then, the distances from each series to the shapelets are computed and the shapelet distance matrix shown in Figure \ref{fig:shapelets_stages} is constructed. Finally, the vectors of distances are used as input to the classifier. In  \cite{Lines2012}, the distance between a shapelet of length $l$ and a time series is defined as the minimum Euclidean distance between the shapelet and all the subsequences of the series of length $l$. Shapelet transformation can be used in combination with any classifier and,  in their proposal,  Lines \textit{et al} experimented with 7 classifiers ( C4.5, 1-NN, Na\"\i ve Bayes, Bayesian Network, Random Forest, Rotation Forest and SVMs) and 26  datasets, showing the benefits of the proposed transformation.

Hills \textit{et al} \cite{Hills2014} provided an extension of \cite{Lines2012} that includes a comprehensive evaluation that analyzes the performance of the 7 aforementioned classifiers using the complete series and the ST as input. As such, the authors concluded that the ST gives rise to improvements in classification accuracy in several datasets. In the same line, Bostrom \textit{et al} \cite{Bostrom2014} proposed another shapelet learning strategy (called \textit{binary ST}) and evaluated their ST in conjunction with an ensemble classifier on 85 UCR datasets, showing that it clearly outperforms conventional approaches of time series classification.

To sum up, a summary of the reviewed methods that employ \textit{Local distance features} can be found in Table \ref{tab:summary_shapelets}.

\vspace{0.5cm}
\begin{center}
{\renewcommand{\arraystretch}{1.2} 
\begin{threeparttable}
\captionof{table}{Summary of local distance feature approaches} 
\begin{tabular}[h]{|l |l |l |l |}
\hline
  \textbf{Authors} & \textbf{Features} & \textbf{Classifier} & \textbf{Datasets} \\ \hline
  Lines \textit{et al.}\cite{Lines2012}  & ST & 7 classifiers* & 18 UCR + 8 own \\ 
  Hills \textit{et al.}\cite{Hills2014}  & ST &  7 classifiers* & 17 UCR + 12 own \\ 
  Bostrom \textit{et al.}\cite{Bostrom2016}   & Binary ST & Ensemble & 85 UCR\\ 
 
\hline 
\end{tabular}

\label{tab:summary_shapelets}

 \begin{tablenotes}
      \small
      \centering
      \item * C4.5, 1-NN, Na\"\i ve Bayes, Bayesian Network, Random Forest, Rotation Forest and SVMs
    \end{tablenotes}
\end{threeparttable}
}
\end{center}
\vspace{0.5cm}

\normalsize	

\vspace{0.3cm}
\subsubsection{Embedded features}
\label{sec:4.3}

The methods presented until now within the \textit{Distance features} category employ the distances directly to create feature vectors representing the series, but this is not the only way to use the distances. In the last approximation within this section, the methods using \textit{Embedded features} do not employ the distances directly as the new representation, but instead, they make use of them to obtain a new representation. In particular, the distances are used to isometrically embed  the series into some Euclidean space while preserving the distances. 

The distance embedding approach is not a specific method for time series. In many areas of research, such as empirical sciences, such as psychology or biochemistry, it is common to have (dis)similarities between the input objects and not the objects per se. As such, one may learn directly in the dissimilarity space mentioned in Section \ref{sec:4.1}, or one may try to find some vectors whose distances approximate the given (dis)similarities. If the given dissimilarities come from the Euclidean distance, it is possible to easily find  some vectors that approximate the given distances. This is known in the literature as \textit{metric multidimensional scaling} \cite{Borg1997}. On the contrary, if the distances are not Euclidean (or even not metric), the embedding approach is not straightforward and many works have addressed this issue in the literature  \cite{Paclik2001}\cite{Graepel1999}\cite{Wilson2014}\cite{Jacobs2000}.

In the case of time series, this approach is particularly advantageous since a vector representation of the series is obtained such that the Euclidean distances between these vectors approximate the given time series distances. The main motivation is that many classifiers are implicitly built on Euclidean spaces \cite{Jacobs2000} and  this approach thus aims to bridge the gap between the Euclidean space and elastic distance measures.  However, as it will be seen, the consistent treatment of new test instances is not straightforward and it is an issue to be considered.

\begin{figure}[h]
\begin{center}
\includegraphics[scale=0.6]{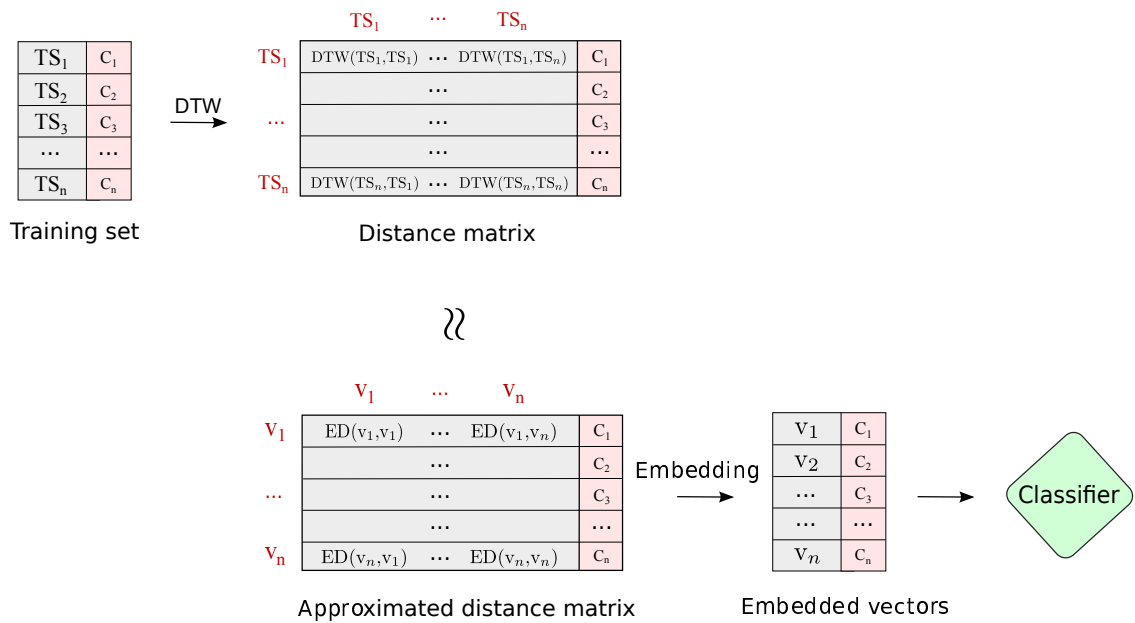}
\caption{Example of the stages of embedded distance features methods using the approach proposed by Hayashi \textit{et al.} \cite{Hayashi2005}.}
\label{fig:embed_stages}
\end{center}
\end{figure}

As examples in TSC, Hayashi \textit{et al.} \cite{Hayashi2005} and Mizuhara \textit{et al.} \cite{Mizuhara2006} proposed, for the first time, a time series embedding approach in which a vector representation of the series is found such that the Euclidean distances between these vectors approximate the DTW distances between the series, as represented in Figure \ref{fig:embed_stages}. They applied three embedding methods: multidimensional scaling, pseudo-Euclidean space embedding and  Euclidean space embedding by the Laplacian eigenmap technique \cite{Belkin2002}. They experimented with linear classifiers and a unique dataset (Australian sign language (ASL)  \cite{Lichman2013}), in which their Laplacian eigenmap-based embedded method achieved a better performance than the 1-NN classifier with DTW.

Another approach presented by  Lei \textit{et al.} \cite{Lei2017} first defines a DTW based similarity measure, called DTWS, following the relation between distances and inner products \cite{Adams2004} (see equation  (\ref{eq:simi})). Then they search for some vectors such that the inner product between these vectors approximates the given DTWS:
\vspace{0.2cm}
\begin{equation}
\label{eq:simi}
DTWS(TS_{i},TS_{j}) = \frac{DTW(TS_{i},0)^{2} + DTW(TS_{j},0)^2 - DTW(TS_{i}, TS_{j})^2}{2}
\vspace{0.2cm}
\end{equation}
where 0 denotes the time series of length one of value 0. Their method learns the optimal vector representation preserving the DTWS by a gradient descent method, but a major drawback is that it learns the transformed time series, but not the transformation itself. The authors propose an interesting solution to deal with the high computational cost of DTW, which consists of assuming that the obtained DTWS similarity matrix is a low-rank matrix. As such, by applying the theory of matrix completion, sampling only $O(n \log n)$ pairs of time series is enough to perfectly  approximate a $n \times n$ low-rank matrix \cite{Sun2016}. However, as mentioned, it is not possible to transform new unlabeled time series, which makes the method rather inapplicable in most contexts.

Finally,  Lods \textit{et al.} \cite{Lods} presented a particular case of embedding that is based on the shapelet transform (ST) presented in the previous section. Their proposal learns a vector representation  of the series (the ST), such that the Euclidean distance between the representations approximates the DTW between the series. In other words, the Euclidean distances between the row vectors representing each series in Figure \ref{fig:shapelets_stages} approximate the DTW distances between the corresponding time series. The main drawback of this approach is the time complexity in the training stage:  first all the DTW distances are computed and then, the optimal shapelets are found by a stochastic gradient descent method. However, once the shapelets are found, the transformation of new unlabeled instances is straightforward, since it is done by computing the Euclidean distance between these series and the shapelets.  Note that they do not use their approach for classifying time series but for clustering, but since it is closely related to the methods in this review and their transformation can be directly applied to classification, it has been included in the taxonomy.

As aforementioned, an important aspect to be considered in the methods using embedded features is the consistent treatment of unlabeled test samples, which depends on the embedding technique used. In the work by Mizuhara \textit{et al.} \cite{Mizuhara2006}, for instance, it is not clearly specified how unlabeled instances are treated.  The method by Lei \textit{et al.} \cite{Lei2017}, on the other hand, learns the transformed data and not the transformation, hence it is not applicable to real problems. Lastly, in the approach by Lods \textit{et al.} \cite{Lods},  new instances are transformed by computing the distance from these new series to the learnt shapelets. 

To end this section, a summary of the reviewed methods employing \textit{Embedded distance features} for TSC can be found in Table \ref{tab:summary_embedded}.

\begin{center}
{\renewcommand{\arraystretch}{1.2} 
\begin{threeparttable}
\captionof{table}{Summary of embedded distance feature approaches} 
\begin{tabular}[h]{|l |l |l |l |}
\hline
  \textbf{Authors} & \textbf{Features} & \textbf{Classifier} & \textbf{Datasets} \\ \hline
  Mizuhara \textit{et al.} \cite{Mizuhara2006} & DTW distance preserving vectors & Linear classifiers & ASL \\
  Lei \textit{et al.} \cite{Lei2017} & DTWS similarity preserving vectors & XGBoost & 6 own\\ 
  Lods \textit{et al.}  \cite{Lods} & DTW distance preserving ST & clustering & 15 UCR \\ 
\hline 
\end{tabular}
\label{tab:summary_embedded}

\end{threeparttable}
}
\end{center}
\vspace{0.5cm}

\subsection{Distance kernels}
\label{sec:5}

The methods within this category do not employ the existing time series distances to obtain a new representation of the series but, instead, they use them to obtain a kernel for time series. Before going in depth into the different approaches, a brief introduction to kernels and kernel methods is presented.

\subsubsection{An introduction to kernels}

The kernel function is the core of kernel methods, a family of pattern recognition algorithms, whose best known instance is the Support Vector Machine (SVM) \cite{Cortes1995}.  Many machine learning algorithms require the data to be in feature vector form,  while kernel methods require only a similarity function -known as kernel- expressing the similarity over pairs of input objects \cite{Shawe-Taylor2004}. The main advantage of this approach is that one can handle any kind of data including vectors, matrices or structured objects, such as sequences or graphs, by defining a suitable kernel which is able to capture the similarity between any two pairs of inputs. The idea behind a kernel is that if two inputs are similar, their output on the kernel will be similar too.

More specifically, a kernel $\kappa$ is a similarity function

\begin{align*}
\kappa : \hspace{0.1cm}  & \mathcal{X} \times \mathcal{X} \rightarrow \mathbb{R} \\ 
   & (x,x')  \rightarrow \kappa(x,x')  
\end{align*}
that for all $x,x' \in \mathcal{X}$ satisfies

\begin{equation}
\label{e}
\kappa(x, x') = \langle \Phi(x) , \Phi(x') \rangle
\end{equation}
where $\Phi$ is the mapping from $\mathcal{X}$ into some high dimensional feature space and $\langle, \rangle$ is an inner product. As equation (\ref{e}) shows, a kernel $\kappa$ is defined by means of a inner product $\langle\,,\rangle$ in some high dimensional feature space. This feature space is called a Hilbert space and the power of kernel methods lies in the implicit use of these spaces \cite{Vapnik1998}.

\begin{figure}[h]
\begin{center}
\includegraphics[scale=0.5]{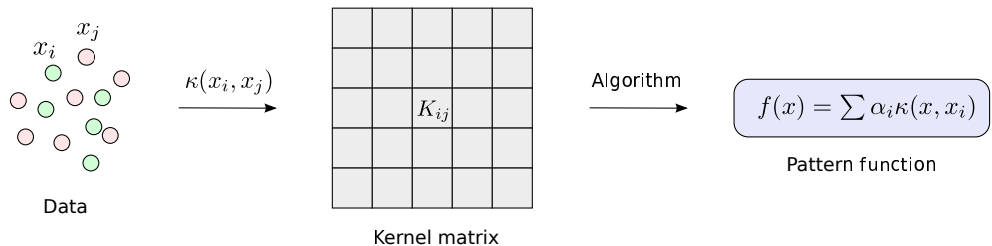}
\caption{The stages involved in the application of kernel methods \cite{Shawe-Taylor2004}.}
\label{fig:kern1}
\end{center}
\end{figure}

In practice, the evaluation of the kernel function is one of the steps within the phases of a kernel method. Figure \ref{fig:kern1} shows the usage of the kernel function within a kernel method and the stages involved in the process. First, the kernel function is applied to the input objects in order to obtain a kernel matrix (also called Gram matrix), which is a similarity matrix with entries $K_{ij}=\kappa(x_{i},x_{j})$ for each input pair $x_{i}$, $x_{j}$. Then, this kernel matrix is used by the kernel method algorithm in order to produce a pattern function that is used to  process unseen instances.

\vspace{0.2cm}

An important aspect to be considered is that the class of similarity functions that satisfies (\ref{e}), and hence they are kernels, coincides with the class of similarity functions that are symmetric and positive semi-definite  \cite{Shawe-Taylor2004}.

\begin{definition} \textbf{(Positive semi-definite kernel)} 
\label{def:psd}
A symmetric function $ \kappa:{\mathcal{X}}\times {\mathcal {X}}\to \mathbb {R} $  satisfying

\begin{equation}
\sum _{i=1}^{n}\sum _{j=1}^{n}c_{i}c_{j}\kappa(x_{i},x_{j})\geq 0
\end{equation}
for any $ n\in \mathbb {N}, x_{1},\dots ,x_{n}\in \mathcal{X}, c_{1},\dots ,c_{n}\in \mathbb{R}$ is called a \textit{positive semi-definite kernel (PSD)} \cite{Scholkopf2002}.
\end{definition}

As such, any PSD similarity function satisfies (\ref{e}) and, since it is a kernel, defines an inner product in some Hilbert space. Moreover, since any kernel guarantees the existence of the mapping implicitly,  an explicit representation for $\Phi$ is not necessary. This is also known as the \textit{kernel trick} (see \cite{Shawe-Taylor2004} for more details). 

\begin{remark} 
We will also refer to a PSD kernel as a  \textit{definite kernel}.
\end{remark}

\begin{remark}
We will informally denominate \textit{indefinite kernels} to non-PSD kernels  which are employed in practice as kernels, even if they do not strictly meet the definition.
\end{remark}

Providing the analytical proof of the positive semi-definiteness of a kernel is rather cumbersome.  In fact, a kernel does not need to have a closed-form analytic expression. In addition, as Figure \ref{fig:kern1} shows, the way of using a kernel function in practice is via the kernel matrix and, hence, the  definiteness of a kernel function is usually evaluated experimentally for a specific set of inputs by analysing the positive semi-definiteness of the kernel matrix.

\begin{definition} \textbf{(Positive semi-definite matrix)} 
A square symmetric matrix $\mathbf{K} \in \mathbb{R}^{n \times n} $  satisfying

\begin{equation}
\label{eq:posi}
\mathbf{v}^T \mathbf{K} \mathbf{v} \geq 0
\end{equation}
for any vector $\mathbf{v} \in \mathbb{R}^{n}$ is called a \textit{positive semi-definite matrix} \cite{Scholkopf2002}. 
\end{definition}

The following well-known result is  obtained from \cite{Shawe-Taylor2004}:
\begin{proposition}
The inequality in equation (\ref{eq:posi}) holds $\Leftrightarrow$ all eigenvalues of $\mathbf{K}$ are non-negative.
\end{proposition}

Therefore, if all the eigenvalues of a kernel matrix are non-negative, this kernel function is considered PSD for the particular instance set in which it has been evaluated. In this manner, the definiteness of a kernel function is usually studied by the eigenvalue analysis of the corresponding kernel matrix. However, a severe drawback of  this approach is that the analysis is only performed for a particular set of instances, and it can not be generalized.

After introducing the basic concepts related to kernels, some examples of different types of kernels are now presented. As previously mentioned, one of the main strengths of kernels is that they can be defined for any type of data, including structured objects, for instance:

\begin{itemize}
\item \textbf{Kernels for vectors:} Given two vectors $\mathbf{x}, \mathbf{x'}$, the popular Gaussian Radial Basis Function (RBF) kernel  \cite{Shawe-Taylor2004} is defined by

\begin{equation}
\label{rbf}
\kappa(\mathbf{x}, \mathbf{x'}) = \exp \left(  - \frac{|| \mathbf{x}- \mathbf{x'} ||^2}{2 \sigma^2}   \right)
\end{equation}
where  $\sigma > 0 $  is a free parameter.
\vspace{0.2cm}

\item \textbf{Kernels for strings:} Given two strings, the $p$-spectrum kernel \cite{Leslie2002} is defined as the number of sub-strings of length $p$ that they have in common. 
\vspace{0.2cm}

\item 
\textbf{Kernels for time series:} Give two time series, a kernel for time series returns a similarity between the series. There are plenty of ways of defining a similarity.  For instance, two time series may be considered similar if they are generated by the same underlying statistical model \cite{Ruping2001}. In this review we will focus on those kernels that employ a time series distance measure to evaluate the similarity between the series.

\end{itemize}

\vspace{0.5cm}
Therefore, in this category denominated \textit{Distance kernels}, instead of using a distance to obtain a new representation of the series, the distances are used to obtain a kernel for time series. As such, the methods in this category aim to take advantage of the potential of time series distances and the power of kernel methods. Two main approaches are distinguished within this category: those that construct and employ an indefinite kernel, and those that construct kernels for time series that are, by definition, PSD.

\vspace{0.3cm}
\subsubsection{Indefinite distance kernels}
\label{sec:9}

The main goal of the methods in this category is to convert a time series distance measure into a kernel. Most distance measures do not trivially lead to PSD kernels, so many works focus on learning with indefinite kernels. The main drawback of learning with indefinite kernels is that the mathematical foundations of the kernel methods are not guaranteed \cite{Ong2004}. The existence of the feature space to which the data is mapped (equation (\ref{e})) is not guaranteed and, thus, due to the missing geometrical interpretation, many good properties of learning in that space (such as orthogonality and projection) are no longer available \cite{Ong2004}.  In addition, some kernel methods do not allow indefinite kernels (due to the implementation or the definition of the method) and some modifications must be carried out, but for others the definiteness is not a requirement. For example, in the case of SVMs, the optimization problem that has to be solved is no longer convex, so reaching the global optimum is not guaranteed \cite{Chen2009}. However, note that good classification results can still be obtained \cite{Bahlmann2002}\cite{Decoste2002}\cite{Shimodaira2006}, so some works focus on studying the theoretical background about SVMs feature space interpretation with indefinite kernels \cite{Haasdonk2005}. Another approach, for instance, employs heuristics on the formulation of SVMs to find a local solution \cite{Chen2006} but, to the best of our knowledge, it has not been applied to time series classification. Converting a distance into a kernel is not a specific challenge of time series and there is a considerable amount of work done in this direction in other contexts \cite{Chen2009}\cite{Haasdonk2004}.

For time series classification, most of the work focuses on employing the distance kernels proposed by Haasdonk \textit{et al.} \cite{Haasdonk2004}. They propose to replace the Euclidean distance in traditional kernel functions, such as the Gaussian kernel in equation \ref{rbf}, by the problem specific distance measure. They called these kernels \textit{distance substitution kernels}. In particular, we will call the following kernel \textit{Gaussian Distance Substitution  (GDS)} \cite{Haasdonk2004}:

\begin{equation}
\label{gdsk}
GDS_{d}(x, x') = \exp \left(  - \frac{d( x, x' )^2}{ \sigma^2}   \right)
\end{equation}
\vspace{0.1cm}

\noindent
where $x, x'$ are two inputs, $d$ is a distance measure and $\sigma > 0 $  is a free parameter. This kernel can be seen as a generalization of the Gaussian RBF kernel presented in the previous section, in which the Euclidean distance is replaced with the distance calculated by $d$. For the GDS kernel, the authors in  \cite{Haasdonk2004} state that GDS$_{d}$ is PSD if and only if $d$ is isometric to an $L$-2 norm, which is generally not the case. As such, the methods which use this type of kernel for time series generally employ indefinite kernels. 

Within the methods employing indefinite kernels, there are different approaches, and for time series classification we have distinguished three main directions (shown in Figure \ref{fig:indefinite_approaches}). Some of them just learn with the indefinite kernels \cite{Kaya2015}\cite{Bahlmann2002}\cite{Shimodaira2006}\cite{Pree2014a}\cite{Jeong2011} using kernel methods that allow this kind of kernels and without taking into consideration that they are indefinite; others argue that the indefiniteness adversely affects the performance and present some alternatives or solutions \cite{Jalalian2013}\cite{Gudmundsson2008}\cite{Chen2015}; finally, others focus on  a better understanding of these distance kernels in order to investigate the reason for the indefiniteness \cite{Zhang2010a}\cite{Lei2007}.

\vspace{1cm}

\begin{center}
\fbox{
  \parbox{8.8cm}{
  \vspace{0.1cm}
   \textit{Indefinite distance kernels}

	\begin{itemize}[label=$\circ$]
	\item Employing indefinite kernels \cite{Jalalian2013}\cite{Gudmundsson2008}\cite{Kaya2015}\cite{Bahlmann2002}\cite{Shimodaira2006}\cite{Pree2014a}\cite{Jeong2011}
	\item Dealing with the indefiniteness   \cite{Jalalian2013}
    \item Regularization \cite{Chen2015} 
	\item Analyzing the indefiniteness  \cite{Zhang2010a}\cite{Lei2007}
	\end{itemize}

  }
}
\vspace{-0.8cm}
 \begin{figure}[H]
\includegraphics[scale=0.00000000001]{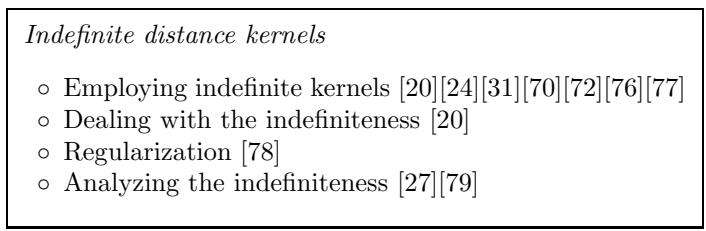}
\caption{Different approaches taken with indefinite distance kernels}
\label{fig:indefinite_approaches}
\end{figure}
\end{center}

\textbf{Employing indefinite kernels}\\

Bahlmann \textit{et al.} \cite{Bahlmann2002} made the first attempt to introduce a time series specific distance measure within a kernel.
They introduced the GDTW measure presented in equation (\ref{gdtw})
as a kernel for character recognition with SVMs. This kernel coincides with the GDS kernel in equation (\ref{gdsk}), in which the distance $d$ is replaced by the DTW distance, i.e., GDS$_ {DTW}$. They remarked that this kernel is not PSD since simple counter-examples can be found in which the kernel matrix has negative eigenvalues. However, they obtained good classification results and argued that for the  UNIPEN\footnotemark \hspace{0.03cm} dataset, most of the eigenvalues of the kernel matrix were measured to be non-negative, concluding that somehow, in the given dataset, the proposed kernel matrix is almost PSD. Following the same direction,  Jeong \textit{et al.} \cite{Jeong2011} proposed a variant of GDS$_ {DTW}$ which employs the Weighted DTW (WDTW)  measure in order to prevent  distortions by outliers, while Kaya \textit{et al.}  \cite{Kaya2015} also employed  the GDS kernel with SVMs, but instead of using the distance calculated by the DTW, they explored other distances derived from different alignment methods of the series, such as Signal Alignment via Genetic Algorithm (SAGA) \cite{Kaya2013}. Pree \textit{et al.} \cite{Pree2014a} proposed a quantitative comparison of different time series similarity measures used either to construct kernels for SVMs or directly for 1-NN classification, concluding that some of the measures benefit from being applied in an SVM, while others do not. Note that in this last work, how they construct the kernel for each distance measure is not exactly detailed. 

There is another method that employs a distance based indefinite kernel but takes a completely different approach to construct the kernel: the idea of this kernel is to, rather than using an existing distance measure, incorporate the concept of alignment between series into the kernel function itself. Many elastic measures for time series deal with the notion of alignment of series. The DTW distance, for instance, finds an optimal alignment between two time series such that the Euclidean distance between the aligned series is minimized. Following the same idea, in  DTAK \cite{Shimodaira2006}, Shimodaira \textit{et al.}  align two series so that their similarity is maximized. In other words, their method finds an alignment between the series that maximizes a given similarity (defined by the user), and this maximal similarity is used directly as a kernel. They give some good properties of the proposed kernel but they remark that it is not PSD, since negative eigenvalues can be found in the kernel matrices of DTAK \cite{Cuturi2011}.

On the other hand,  Gudmundsson \textit{et al.} \cite{Gudmundsson2008} employed the DTW based similarity measures they proposed (shown in equantion (\ref{gdtw})) directly as kernels. Their method achieved low classification results and the authors claimed that another way of introducing a distance into a SVM is by using the distance features introduced in Section \ref{sec:4.1}. They compared the performance of DTW based distance features and DTW based distance kernels, concluding that distance features outperform the distance kernels due to the indefiniteness of these second ones. \\

\footnotetext{On-line handwritten digit data set \cite{Guyon1994}}

%

\hspace{0.5cm}\textbf{Dealing with the indefiniteness}\\

There is a group of methods that attribute the poor performance of their kernel methods to the indefiniteness, and propose some alternatives or solutions to overcome these limitations. Jalalian \textit{et al.} \cite{Jalalian2013}, for instance, proposed the use of a special SVM called Potential Support Vector Machine (P-SVM)\cite{Hochreiter2006} to overcome the shortcomings of learning with indefinite kernels. They employed the GDS$_ {DTW}$ kernel within this SVM classifier which is able to handle kernel matrices that are neither positive definite nor square. They carried out an extensive experimentation including a comparison of their method with the 1-NN classifier and with the methods presented in \cite{Gudmundsson2008}.  They conclude that their  DTW based P-SVM method significantly outperforms both distance features and indefinite distance kernels, as well as the benchmark methods in 20 UCR datasets. \\

\hspace{0.5cm}\textbf{Regularization}\\

Another approach that tries to  overcome the use of indefinite kernels consists of regularizing the indefinite kernel matrices to obtain  PSD matrices.  As previously mentioned, a matrix is PSD if and only if all its eigenvalues are non-negative,  and a kernel matrix therefore can be regularized by clipping all the negative eigenvalues to zero, for instance. This technique has been usually applied for non-temporal data \cite{Chen2009}\cite{Wu2005}\cite{Wu2005a} but it is rather unexplored in the domain of indefinite time series kernels. Chen \textit{et al.} \cite{Chen2015} proposed a Kernel Sparse Representation based Classifier (SRC) \cite{Zhang2012} with some indefinite time series kernels and applied spectrum regularization to the kernel matrices. In particular, they employed the GDS$_{DTW}$, GDS$_{ERP}$ (Edit distance with Real Penalty (ERP) \cite{Chen2004}) and GDS$_{TWED}$  (Time Warp Edit Distance (TWED) \cite{Marteau2009}) kernels and their method checks whether the kernel matrix obtained for a specific dataset is PSD. If it is not, the corresponding kernel matrix is regularized using the spectrum clip approach. 

Regarding this approach, it is also worth mentioning that in the work by Gudmundsson \textit{et al.} \cite{Gudmundsson2008}, the authors point out that they tried to apply some regularization to the kernel matrix subtracting the smallest eigenvalue from the diagonal but they found out that the method achieved a considerably low performance. Additionally, the authors added that matrix regularization can lead to matrices with large diagonal entries, which may result in overfitting  \cite{Weston2003}.

Finally, the consistent treatment of training and new unlabeled instances  is not straightforward and  is also a matter to bear in mind \cite{Chen2009}. When new unlabeled instances arrive, the kernel between them and the training set has to be computed. If the kernel matrix corresponding to the training set has been regularized, the kernel matrix corresponding to the unlabeled set should also be modified in a consistent way, which is not a trivial operation. Therefore, the benefit of matrix regularization in the context of time series is an open question.\\

\textbf{Analyzing the indefiniteness}\\

The last group of methods do not focus on solving the problems of learning with indefinite kernels but, instead, they focus on a better understanding of these distance kernels and their indefiniteness.  Lei \textit{et al.} \cite{Lei2007} theoretically analyze  the GDS$_ {DTW}$ kernel, proving that it is not a PSD kernel. This is because DTW is not a metric (it violates the triangle inequality \cite{Casacuberta1987}) and non-metricity prevents definiteness \cite{Haasdonk2004}. That is, if $d$ is not metric,  GDS$_ {d}$ is not PSD.  However, the contrary is not true and, hence, the metric property of a distance measure is not a sufficient condition to guarantee a PSD kernel. In any case, Zhang \textit{et al.} \cite{Zhang2010a}, hypothesized that kernels based on metrics give rise to better performances than kernels based on distance measures which are not metrics. As such, they define what they called the Gaussian Elastic Metric Kernel (GEMK), a family of GDS kernels in which the distance $d$  is replaced by an elastic measure which is also a metric. They employed GDS$_{ERP}$ and GDS$_{TWED}$ and stated that, even if the  definiteness of these kernels is not guaranteed,  they did not observe any violations of their definiteness in their experimentation on 20 UCR datasets. In fact, these kernels are shown to perform better than the GDS$_ {DTW}$ and the Gaussian kernel in those experiments. The authors attribute this to the fact that the proposed measures are both elastic and obey metricity. In order to provide some information about the most common distance measures applied in this context, table \ref{tab:elastic_metric} shows a summary of properties of the main distance measures employed in this review. In particular, we specify if a given distance measure $d$ is a metric or not, if it is an elastic measure or not, and if the corresponding GDS$_ {d}$ is proven to be PSD or not. 

To sum up, there are some results that suggest a relationship between the metricity of the distance and the performance of the corresponding distance kernel. However, it is hard to investigate the contribution of metricity in the accuracy since several factors take part in the classification task. The definiteness of a distance kernel seems to be related to the metricity of given distance -metric distances seem to lead to kernels that are closer to definiteness than those based on non-metric distances-, and the definiteness of a kernel may directly affect on the accuracy. In short, the relationship between metricity, definiteness and performance is not clear and is, thus, an interesting future direction of research.

\begin{center}
{\renewcommand{\arraystretch}{1.2} 
\captionof{table}{Summary of distance properties used in GDS} 
\begin{tabular}[h]{|r |r |r | r| }
\hline
 \textbf{Distance} & \textbf{metric} & \textbf{elastic} &  $GDS_{d}$ \textbf{is PSD}\\ \hline
  Euclidean      &   \checkmark   & $\times$ & \checkmark \\ 
   DTW           &  $\times$      & \checkmark & $\times$\\ 
   ERP           &  \checkmark    & \checkmark & $\times$\\ 
   TWED           &  \checkmark    & \checkmark & $\times$\\ 
\hline
\end{tabular} 
\label{tab:elastic_metric}
}
\end{center}
\vspace{0.5cm}

To conclude, a summary of the reviewed methods of \textit{Indefinite distance kernels}  can be found in Table \ref{tab:summary_GDS}.

{\renewcommand{\arraystretch}{1.2} 
\captionof{table}{Summary of indefinite kernel approaches} 
\hspace{-1cm}\begin{tabular}[h]{|l |l |l |l |}
\hline
 \textbf{Authors} & \textbf{Kernel} & \textbf{Classifier} & \textbf{Datasets} \\ \hline
 \textbf{Employing indefinite kernels} & & & \\
 \hspace{0.5cm} Bahlmann \textit{et al.}  \cite{Bahlmann2002}    & GDS$_{DTW}$ & SVMs & 1 (UNIPEN)\\ 
  \hspace{0.5cm} Jeong \textit{et al.} \cite{Jeong2011}	    & GDS$_{WDTW}$ & SVDD\footnotemark, SVMs & 20 UCR  \\ 
  \hspace{0.5cm} Kaya \textit{et al.}   \cite{Kaya2015} & GDS +  alignment based distances & SVMs  & 40  UCR   \\ 
  \hspace{0.5cm} Pree \textit{et al.}\cite{Pree2014a} & Unespecified similarity based kernels & SVMs  & 20  UCR \\
  \hspace{0.5cm} Shimodaira \textit{et al.} \cite{Shimodaira2006} & DTAK  &  SVMs & ATR  \\
   \hspace{0.5cm} Gudmundsson \textit{et al.}  \cite{Gudmundsson2008}  & NDTW, GDS$_{DTW}$& SVMs &  20 UCR \\ 

\textbf{Dealing with the indefiniteness} & & & \\ 
 \hspace{0.5cm} Jalalian \textit{et al.} \cite{Jalalian2013} & GDS$_{DTW}$ & P-SVM &  20 UCR \\ 
\textbf{Regularization} & & & \\  
  \hspace{0.5cm} Chen \textit{et al.}\cite{Chen2015}  & GDS$_{DTW}$, GDS$_{ERP}$ , GDS$_{TWED}$  &  KSRC\footnotemark & 16 UCR\\ 
 \textbf{Analyzing the indefiniteness} & & & \\ 
  \hspace{0.5cm} Lei \textit{et al.} \cite{Lei2007}  & GDS$_{DTW}$ & SVMs & 4 UCR  \\ 
  \hspace{0.5cm} Zhang \textit{et al.}	 \cite{Zhang2010a}   & GDS$_{ERP}$, GDS$_{TWED}$ & SVMs  &  20 UCR  \\

\hline 
\end{tabular}

\label{tab:summary_GDS}

} 
\footnotetext{Support Vector Data Descriptor \cite{Hochreiter2006}\cite{Tax2004}}
\footnotetext{Kernel Sparse Representation based Classifiers  \cite{Zhang2012}}

\vspace{0.3cm}
\subsubsection{Definite distance kernels}
\label{sec:10}

We have included in this section those methods that construct distance kernels for time series which are, by definition, PSD. First of all, we want to remark that there are other kernels for time series in the literature that are PSD  but have not been included in this review. The reason is that we have only incorporated those kernels that are based on time series distances and, in particular, those which construct the kernel functions directly on the raw series. Conversely, the Fourier kernel \cite{Ruping2001} computes the inner product of the Fourier expansion of two time series, and hence, does not compute the kernel on the raw series but on the Fourier expansion of them.  Another example is the kernel by Gaidon \textit{et al.} \cite{Gaidon2011} for action recognition, in which the kernel is constructed on the auto-correlation of the series. There are also smoothing kernels that smooth the series with different techniques and then define the kernel for those smoothed representations \cite{Troncoso2015}\cite{Kumara2008}\cite{Sivaramakrishnan2004}\cite{Lu2008}. On the contrary, we will focus on those that define a kernel directly on the raw series. Regarding those included, all of them  aim to introduce the concept of time elasticity directly within the kernel function by means of a distance, and we can distinguish two main approaches: in the first, the concept of the alignment between series is exploited, while in the second, the direct construction of PSD kernels departing from a given distance measure is addressed.

Xue \textit{et al.} \cite{Xue2017} proposed the Altered Gaussian DTW (AGDTW) kernel, in which, first, the alignment that minimizes the Euclidean distance between the series is found, as in DTW. For each pair of time series ${TS_{i}}$ and ${TS_{j}}$, once this alignment is found, the series are modified respect to this alignment resulting in ${TS_{i}}'$ and ${TS_{j}}'$. Then, if  $S$ is the maximum length of both series, the AGDTW kernel is defined as follows:

$$ \kappa_{AGDTW}(TS_{i}, TS_{j}) = \sum_{s=1}^{S} \exp \left(- \frac{|| {TS_{i}}_s'- {TS_{j}}_s' ||^2}{\sigma^2} \right)$$
Since AGDTW is, indeed, a sum of Gaussian kernels, they provide the proof of the definiteness of the proposed kernel.

There is another family of methods that also exploits the concept of alignment but, instead of considering just the optimal one, considers the sum of the scores obtained by all the possible alignments between the two series. Cuturi \textit{et al.} \cite{Cuturi2007} claimed that two series can be considered similar not only if they have one single good alignment, but rather if they have several good alignments. They  proposed the Global Alignment (GA) kernel that takes into consideration  all the alignments   between the series and provide the proof of its positive definiteness under certain mild conditions.  It is worth mentioning that they obtain kernel matrices that are exceedingly diagonally dominant, that is, that the values of the diagonal in the matrix are many orders of magnitude larger than those out of the diagonal. Thus, they use the logarithm of the kernel matrix because of possible numerical problems. That transformation makes the kernel indefinite (even if it is not  indefinite per se), so they apply some kernel regularization to turn all its eigenvalues positive. However, since the kernel they obtain is PSD and it is because of the logarithm transformation that it becomes indefinite, it has been included within this section. In \cite{Cuturi2011}, Cuturi \textit{et al.} elaborate on the GA kernels, give some theoretical insights and introduce an extension called Triangular Global Alignment (TGA) kernel, which is faster to compute and also PSD.

There is another kernel that takes a similar approach:  Wachman \textit{et al.} \cite{Wachman2009}, in their work about periodic time series in astronomy, investigate the similarity between just shifted time series. In this way, they define a kernel that takes into consideration the contribution of all possible alignments obtained by employing just time shifting: 

$$ K_{shift}(TS_{i},TS_{j}) = \sum_{s=1}^{n} e^{\gamma \langle TS{_{i}}, TS{_{j}}_{+ s}\rangle} $$
where $\gamma\geq0$ is a user-defined constant.  In this way, the kernel is defined by means of a sum of inner products between $TS_{i}$ and all the possible shifted versions of $TS_{j}$ with a shift of $s$ positions. The authors provided the proof of the PSD of the proposed kernel.

On the other hand,  there are methods that, instead of focusing on alignments, address the construction of PSD kernels departing from a given distance measure. These methods can be seen as refined versions of the GDS kernel in which the obtained kernel is PSD.  Marteau \textit{et al.} \cite{Marteau2010}  elaborate on the indefiniteness of GDS kernels derived from  elastic measures, even when such measures are metrics. As previously mentioned, metricity is not a sufficient condition to obtain PSD kernels. They postulated that elastic measures do not lead to PSD kernels due to the presence of $min$ or $max$ operators in their definitions, and define a kernel where they replaced the  $min$ or $max$ operators by a sum ($\sum$). In \cite{Marteau2012}, these same authors define what they called an elastic inner product, $eip$. Their goal was to embed the time series into an inner product space that somehow generalizes the notion of the Euclidean space, but retains the concept of elasticity. They provide the proof of the existence of such a space and showed  that this $eip$ is, indeed, a PSD kernel. Since any inner product induces a distance \cite{Greub1975}, they obtained a new elastic metric distance  $\delta_{eip}$ that avoids the use of $min$ or $max$ operators. They evaluated the obtained distance within a SVM  by means of the GDS$_{\delta_{eip}}$ kernel, in order to compare the performance of $\delta_{eip}$ with the Euclidean and DTW measures. Their experimentation showed that elastic inner products can bring a significant  improvement in accuracy compared to the Euclidean distance, but the  GDS$_{DTW}$ kernel outperforms the proposed GDS$_{\delta_{eip}}$ in the majority of the datasets.

They extended their work in \cite{Marteau2014} and introduced the Recursive Edit Distance Kernels (REDK), a method to construct PSD kernels departing from classical edit or time-warp distances. The main procedure to obtain PSD kernels is, as in the previous method, to replace the $min$ or $max$ operators by a sum. They provided the proof of the definiteness of these kernels when some simple conditions are satisfied, which are weaker than those proposed in \cite{Cuturi2007} and  are satisfied by any classical elastic distance defined by a recursive equation. Note that, while in \cite{Marteau2012} the authors define an elastic distance and construct PSD kernels with it, in \cite{Marteau2014} the authors present a method to construct a PSD kernel departing from any existing elastic distance measure. In this manner, the REDK can be seen as a refined version of the GDS kernel which leads to PSD kernels. In this manner, they proposed the REDK$_{DTW}$, REDK$_{ERP}$ and REDK$_{TWED}$ methods and compare their performance with the corresponding distance substitutions kernels GDS$_{DTW}$, GDS$_{ERP}$ and GDS$_{TWED}$. An interesting result they reported is that REDK methods seem to improve the performance of non-metric measures in particular. That is, while the accuracies of  REDK$_{ERP}$ and REDK$_{TWED}$ are slightly better than the accuracies of GDS$_{ERP}$ and GDS$_{TWED}$, in the case of DTW the improvement is really significant. In fact, they presented some measures to quantify the deviation from definiteness of a matrix and showed that while GDS$_{ERP}$ and GDS$_{TWED}$ are \textit{almost definite}, GDS$_{DTW}$ is rather far from being definite. This makes us wonder again if metricity implies proximity to definiteness, and in addition, if accuracy is directly correlated to the definiteness of the kernel. 
 
Furthermore, they explored the possible impact of the indefiniteness of the kernels on the accuracy by defining several measures to quantify the deviation from definiteness based on eigenvalue analysis. If  $\mathbf{D}_{\delta}$ is a distance matrix,  GDS$_{\mathbf{D}_{\delta}}$ is PSD if and only if $\mathbf{D}_{\delta}$ is negative definite \cite{Cortes2004}, and $\mathbf{D}_{\delta}$ is negative definite if it has a single positive eigenvalue. In this manner, the authors studied the deviation from definiteness of some distance matrices, and stated that when the distance matrix $\mathbf{D}_{\delta}$ was far from being negative definite, the REDK$_{\delta}$  outperforms  the GDS$_{\delta}$ kernel in general, while when the matrix is close to negative definiteness, REDK$_{\delta}$ and GDS$_{\delta}$ perform similarly.

Recently, Wu \textit{et al.} \cite{Wu2018} introduced another distance substitution kernel, called D2KE, that addresses the construction of a family of PSD kernels departing from any distance measure.  It is not specific for time series but in their experimentation they include a kernel for time series departing from the DTW distance measure. Their kernel employs a probability distribution over random structured objects (time series in this case) and defines a kernel that takes into account the distance from two series to the randomly sampled objects. In this manner, the authors point out that the D2KE kernel can be interpreted as a soft version of the GDS kernel, which is PSD. Their experimentation on 4 time series datasets showed that their D2KE$_{DTW}$ kernel outperforms other distance based approaches such as 1-NN or GDS$_{DTW}$   both in accuracy and computational time.

To conclude this section, a summary of the reviewed methods on \textit{Definite distance kernels} can be found in Table \ref{tab:summary_elastic}.

\vspace{0.5cm}
{\renewcommand{\arraystretch}{1.2} 
\begin{center}
\begin{threeparttable}[h]
\captionof{table}{Summary of definite distance kernels} 
\begin{tabular}{|l |l |l |l |}
\hline
    \textbf{Authors} & \textbf{Kernel} & \textbf{Classifier} & \textbf{Datasets}  \\ \hline
  Xue \textit{et al.}\cite{Xue2017} & AGDTW & KSRC, SVMs & 4 UCR  \\ 
 
  Cuturi \textit{et al.}   \cite{Cuturi2007} & GA  & SVMs & TI46\footnotemark  \\ 
  
  Cuturi \textit{et al.} \cite{Cuturi2011}  & TGA  &SVMs & 5 UCI   \\ 

  Marteau \textit{et al.} \cite{Marteau2012}  & GDS$_{\delta_{eip}}$   &        SVMs & 20 UCR   \\ 
  Marteau \textit{et al.} 
 \cite{Marteau2014}  & \tiny{REDK$_{DTW}$, REDK$_{ERP}$, REDK$_{TWED}$ }  &   SVMs & 20 UCR  \\ 

  Wu \textit{et al.} 
 \cite{Wu2018}  & D2KE &   SVMs & 3 UCI + 1 own \\ 
 
 Wachman \textit{et al.} \cite{Wachman2009} & K$_{shift}$ & SVMs & Astronomy \\

\hline 
\end{tabular}
\label{tab:summary_elastic}
\end{threeparttable}
\end{center}
\vspace{0.5cm}
\footnotetext{TI46 \cite{Liberman1993} speech dataset}
}

\section{Discussion and future work}
\label{sec:6}

In this paper, we have presented a review on distance based time series classification  and  have included a taxonomy that categorizes all the discussed methods depending on how each approach uses the given distance. We have seen that, from the most general point of view, there are three main approaches: those that directly employ the distance together with the 1-NN classifier, those that use the distance to obtain a new feature representation of the series, and those which construct kernels for time series departing from distance measure. The first approach has been widely reviewed, so we refer the reader to \cite{Wang2013}\cite{Ding2008}\cite{Serra2014} for more details about the discussion.

Regarding the methods that employ a distance to obtain a new feature representation of the series, these approaches have been considerably studied for time series as it bridges the gap between traditional classifiers -that expect a vector as input- and time series data, taking advantage of the  existing time series distances. In addition, some methods within this category have outperformed existing time series benchmark classification methods \cite{Kate2015}. Note that distance features can be seen as a preprocessing step, where a new representation of the series is found which is independent of the classifier. Depending on the specific problem, these representations vary and can be more discriminative and appropriate than the original raw series \cite{Hills2014}. As such, an interesting point that has not been addressed yet is to compare the different transformations of the series in terms of how discriminative they are for classification. 

Nevertheless, learning with the distance features can often become cumbersome depending on the size of the training set and a dimensionality reduction technique must be applied in many cases in order to lower the otherwise intractable computational cost. Some of the methods \cite{Iwana2017}\cite{Goebel2015} reduce the dimensionality of the distance matrix once it is computed. Another direction focuses on time series prototype selection \cite{Iwana2017}, that is, selecting some representative time series in order to compute only the distances to them instead of to the whole training set. It is worth mentioning that there has been some work done in this context in other dissimilarity based learning problems \cite{Pekalska2006} but it is almost unexplored in TSC. Due to the interpretability of the time series and, in particular, of their prototypes, we believe that this is a promising future direction of research. 

Another feature  based method consists of embedding. The embedded distance features have only been  employed in combination with linear classifiers \cite{Mizuhara2006} or the tree based XGBoost classifier \cite{Lods}, which, in our opinion, do not take direct advantage of the transformation.  The main idea of the embedded features is that if the Euclidean distances of the obtained features are computed, the original time series distances are approximated. In this way, we believe that  classifiers that compute Euclidean distances within the classification task (such as the SVM with the RBF kernel, for instance) will profit better from this representation. In addition, in the particular case of kernel methods, the use of embedded features can be seen as a kind of regularization; the RBF kernel obtained from the embedded features would be a definite kernel that approximates the GDS indefinite kernel.

As already pointed out, the third way of using a distance measure is trying to construct kernels departing from these existing distances. However, these distances do not generally lead to PSD kernels. Both distance features and distance kernel approaches are not  specific for time series, and some work has been done to compare the benefits of each approach in a general context. Chen \textit{et al.} \cite{Chen2009} mathematically studied  the influence of distances features and distances kernels within SVMs in a general framework. In time series classification, Gudmundsson \textit{et al.} \cite{Gudmundsson2008} and Jalalian \textit{et al.} \cite{Jalalian2013} address the problem of experimentally evaluating  whether it is preferable to use distance features or distance kernels. Both works assert that the indefiniteness of the distance kernels negatively affects  the performance, although their proposals are restricted to the DTW distance. It would be interesting to comprehensively compare these two approaches taking into account different distances, kernels and classifiers in order to draw more general conclusions.

The problem of the definiteness of a kernel has been widely addressed within the methods in this review. Note that the definiteness of a kernel guarantees the mathematical foundations of the kernel method and, therefore, it seems natural to think that definiteness and performance are correlated -which is the assumption of almost all the methods. Some authors confirm that the performance is still good and do not care about the indefiniteness of the kernels, while, in general, the research focuses mainly on trying to somehow deal with the indefiniteness of the kernels. Isolating the contribution of the definiteness of a kernel to the performance is rather challenging due to the many other factors (optimization algorithm or the choice of the kernel function) that also affect it. However, since the relation between definiteness and accuracy is a general matter -not specific for time series, and in fact, not specific for distance kernels-, a promising future direction would be to evaluate whether there exists or not a direct correlation between them.

Within the methods that try to deal with the indefiniteness there are two main directions. The first  uses kernel based classifiers that can handle indefinite kernels.  This approach is  almost unexplored  in time series classification, since only the P-SVM by Jalalian \textit{et al.} \cite{Jalalian2013} has been applied, achieving very competitive results. Indeed, there are some studies on learning with indefinite kernels  from a general point of view \cite{Ong2004}, and considering that indefinite kernels appear often within TSC, this approach may be interesting future work.

The second approach, called kernel regularization, aims at adapting the indefinite kernel to be PSD. As in the previous direction, this is also an almost unexplored approach for time series. Only eigenvalue analysis has been applied with ambiguous results. Chen \textit{et al.} \cite{Chen2015a} used eigenvalue regularization techniques but they do not evaluate the regularization itself, while Gudmundsson \textit{et al.} \cite{Gudmundsson2008} argued that the method after kernel regularization achieves lower performance than the method with the indefinite kernel. One of the main shortcomings of this specific regularization is that it is data dependent, and, in addition, the consistent treatment of new test samples is not straightforward. As mentioned previously, it is not clear whether regularization is helpful or whether the new kernel becomes so different from the initial one that the information loss is too big; this is an open question which has not been studied in detail.

As aforementioned, another direction focuses on a better understanding of the indefiniteness of these kernels. Concerning the GDS kernels, which are, indeed, distance kernels which are valid for any type of data, the first attempt in the time series domain was to define kernels departing from distances that are metrics.  Although it has been proven that the metric property does not guarantee the definiteness of the induced GDS kernel, Zhang \textit{et al.} \cite{Zhang2010a} argued that the performance of metric distance  kernels is significantly better than those defined with non-metric distances, suggesting that kernels with metric distances are closer to definiteness.  In addition, Marteau \textit{et al.} \cite{Marteau2014}  conjecture that the reason of the indefiniteness is the presence of $min$ or $max$ operators in the recursive definition of time series distance measures. An interesting observation is that these discussions arise from time series distances but are, indeed, general issues concerning the  characteristics of a distance measure and the derived GDS kernel. Even if the mentioned works address the relation between metricity and definiteness, this connection is not still clear. Due also to the generalizability of the problem and the possible applications, it is an interesting future research direction. 

Cuturi \textit{et al.} \cite{Cuturi2007}, by contrast, focused on the specific challenge of constructing ad-hoc kernels for time series. As such, they found  a direct way of constructing PSD kernels that take into account the time elasticity by defining a kernel that does not consider just the optimal alignment between two series but, instead, it considers all the possible alignments.  Moreover, given an elastic distance measure defined by a recursive equation, Marteau \textit{et al.} \cite{Marteau2009} address the construction of distance based PSD kernels. Their kernel can be seen as a particular case of GDS kernel for elastic measures that become PSD by replacing the $min$ or $max$ operators in the  recursive definition of the distance by a sum. By using this \textit{trick}, they obtain kernels for time series that take into account time elasticity and are also PSD.  Their comprehensive experimentation shows that SVM based approaches which use these kernels clearly outperform the 1-NN benchmark approaches, even for the DTW distance. Furthermore, they reported that the REDK kernel brings significant improvement in comparison with the GDS kernel, especially when the kernel matrices of the GDS kernels are far from definiteness, which in their particular case corresponds to the non-metric measures. However, they experimented  with just two metric and one non-metric measures which is not enough to draw strong conclusions. 

It is also worth mentioning that many methods introduced in the taxonomy are not specific for time series, but become specific when a time series distance is employed. In particular, only the methods that are based on shapelets and the methods that construct kernels for time series considering the concept of alignment between series are specific for time series. The rest of the methods are general methods of distance based classification for any type of data. An interesting observation is that questions or problems arising for time series can be extrapolated to a general framework. In the same manner, some of the presented approaches are specific for some classifiers (1-NN, kernel methods), while others can be used in combination with any classifier. Note also that many of the methods are directly applicable in the case of multivariate or streaming time series, provided a suitable distance for these kind of data.

To conclude, note that in contrast to the number and variety of existing kernels for other types of data, there are rather few benchmark kernels for time series in the literature \cite{Shawe-Taylor2004}. Therefore, we would like to highlight the value of these kernels for time series, especially those that are able to deal with the temporal nature of the series and are PSD.  

\vspace{1cm}

\begin{acknowledgements}

\vspace{0.5cm}

This research is supported by the Basque Government through the BERC 2018-2021 program and by Spanish Ministry of Economy and Competitiveness MINECO through BCAM Severo Ochoa excellence accreditation SEV-2013-0323 and through project TIN2017-82626-R funded by (AEI/FEDER, UE) and acronym “GECECPAST”. In addition, by the Research Groups 2013-2018 (IT-609-13) programs (Basque Government), TIN2016-78365-R (Spanish Ministry of Economy, Industry and Competitiveness).  A. Abanda is also supported by the grant BES-2016-076890.

\end{acknowledgements}


\begin{thebibliography}{100}

\bibitem{Keogh2002}
Eamonn Keogh and Shruti Kasetty.
\newblock {On the need for time series data mining benchmarks}.
\newblock {\em Proceedings of the 8th ACM SIGKDD International Conference on
  Knowledge Discovery and Data Mining}, page 102, 2002.

\bibitem{Esling2012}
Philippe Esling and Carlos Agon.
\newblock {Time-series data mining}.
\newblock {\em ACM Computing Surveys}, 45(1):1--34, 2012.

\bibitem{Bagnall2017}
Anthony Bagnall, Jason Lines, Aaron Bostrom, James Large, and Eamonn Keogh.
\newblock {The great time series classification bake off: a review and
  experimental evaluation of recent algorithmic advances}.
\newblock {\em Data Mining and Knowledge Discovery}, 31(3):606--660, 2017.

\bibitem{Fu2011}
Tak~Chung Fu.
\newblock {A review on time series data mining}.
\newblock {\em Engineering Applications of Artificial Intelligence},
  24(1):164--181, 2011.

\bibitem{Xing2010}
Zhengzheng Xing, Jian Pei, and Eamonn Keogh.
\newblock {A brief survey on sequence classification}.
\newblock {\em ACM SIGKDD Explorations Newsletter}, 12(1):40, 2010.

\bibitem{ChristosFaloutsosM.Ranganathan1994}
Christos Faloutsos, M.~Ranganathan, and Yannis Manolopoulos.
\newblock {Fast subsequence matching in time-series databases}.
\newblock {\em ACM SIGMOD International Conference on Management of Data},
  pages 419--429, 1994.

\bibitem{Popivanov2002}
I.~Popivanov and R.~J. Miller.
\newblock {Similarity Search Over Time-Series Data Using Wavelets}.
\newblock {\em Proceedings 18th International Conference on Data Engineering
  (ICDE)}, pages 212--221, 2002.

\bibitem{FlipKornH.V.Jagaciish1997}
Flip Korn, H.~V. Jagaciish, and Christos Faloutsos.
\newblock {Efficiently Supporting Ad Hoc Queries Sequences in Large Datasets of
  Time for Systems}.
\newblock {\em Proceedings of the 1997 ACM SIGMOD International Conference on
  Management of Data}, pages 289--300, 1997.

\bibitem{Bagnall2014}
Anthony Bagnall and {Gareth Janacek}.
\newblock {A run length transformation for discriminating between auto
  regressive time series}.
\newblock {\em Journal of Classification}, 31(October):274--295, 2014.

\bibitem{Corduas2008}
Marcella Corduas and Domenico Piccolo.
\newblock {Time series clustering and classification by the autoregressive
  metric}.
\newblock {\em Computational Statistics and Data Analysis}, 52(4):1860--1872,
  2008.

\bibitem{Smyth1997}
Padhraic Smyth.
\newblock {Clustering sequences with hidden Markov models}.
\newblock {\em Advances in Neural Information Processing Systems}, 9:648--654,
  1997.

\bibitem{Berndt1994}
Donald Berndt and James Clifford.
\newblock {Using dynamic time warping to find patterns in time series}.
\newblock {\em Workshop on Knowledge Knowledge Discovery in Databases},
  398:359--370, 1994.

\bibitem{Wang2013}
Xiaoyue Wang, Abdullah Mueen, Hui Ding, Goce Trajcevski, Peter Scheuermann, and
  Eamonn Keogh.
\newblock {Experimental comparison of representation methods and distance
  measures for time series data}.
\newblock {\em Data Mining and Knowledge Discovery}, 26(2):275--309, 2013.

\bibitem{Chen2013}
Yanping Chen, Bing Hu, Eamonn Keogh, and Gustavo~E.A.P.A. Batista.
\newblock {DTW-D: Time Series Semi-Supervised Learning from a Single Example}.
\newblock {\em Proceedings of the 19th ACM SIGKDD International Conference on
  Knowledge Discovery and Data Mining}, page 383, 2013.

\bibitem{Ding2008}
Hui Ding, Goce Trajcevski, Peter Scheuermann, Xiaoyue Wang, and Eamonn Keogh.
\newblock {Querying and mining of time series data: experimental comparison of
  representations and distance measures}.
\newblock {\em Proceedings of the VLDB Very Large Database Endowment},
  1(2):1542--1552, 2008.

\bibitem{Lines2015}
Jason Lines and Anthony Bagnall.
\newblock {Time series classification with ensembles of elastic distance
  measures}.
\newblock {\em Data Mining and Knowledge Discovery}, 29(3):565--592, 2015.

\bibitem{Xi2006}
Xiaopeng Xi, Eamonn Keogh, Christian Shelton, Li~Wei, and Chotirat~Ann
  Ratanamahatana.
\newblock {Fast time series classification using numerosity reduction}.
\newblock {\em Proceedings of the 23rd ICML International Conference on Machine
  learning}, pages 1033----1040, 2006.

\bibitem{Tan2005}
Pang-Ning Tan, Michael Steinbach, and Vipin Kumar.
\newblock {\em {Introduction to Data Mining}}.
\newblock Addison-Wesley, Reading, MA, addison-we edition, 2005.

\bibitem{Kate2015}
Rohit~J. Kate.
\newblock {Using dynamic time warping distances as features for improved time
  series classification}.
\newblock {\em Data Mining and Knowledge Discovery}, 30(2):283--312, 2015.

\bibitem{Jalalian2013}
Arash Jalalian and Stephan~K. Chalup.
\newblock {GDTW-P-SVMs: Variable-length time series analysis using support
  vector machines}.
\newblock {\em Neurocomputing}, 99:270--282, 2013.

\bibitem{Marteau2014}
Pierre-Fran{\c{c}}ois Marteau and Sylvie Gibet.
\newblock {On Recursive Edit Distance Kernels With Application to Time Series
  Classification}.
\newblock {\em IEEE Transactions on Neural Networks and Learning Systems},
  26(6):1--15, 2014.

\bibitem{Iwana2017}
Brian~Kenji Iwana, Volkmar Frinken, Kaspar Riesen, and Seiichi Uchida.
\newblock {Efficient temporal pattern recognition by means of dissimilarity
  space embedding with discriminative prototypes}.
\newblock {\em Pattern Recognition}, 64(September 2016):268--276, 2017.

\bibitem{Hills2014}
Jon Hills, Jason Lines, Edgaras Baranauskas, James Mapp, and Anthony Bagnall.
\newblock {Classification of time series by shapelet transformation}.
\newblock {\em Data Mining and Knowledge Discovery}, 28(4):851--881, 2014.

\bibitem{Gudmundsson2008}
S~Gudmundsson, T~P Runarsson, and S~Sigurdsson.
\newblock {Support vector machines and dynamic time warping for time series}.
\newblock In {\em Joint Conference on Neural Networks (IEEE World Congress on
  Computational Intelligence)}, pages 2772--2776, 2008.

\bibitem{Cuturi2007}
Marco Cuturi and Jp~Vert.
\newblock {A kernel for time series based on global alignments}.
\newblock {\em IEEE Transactions on Acoustics, Speech and Signal Processing},
  1:413--416, 2007.

\bibitem{Jeong2015}
Young-Seon Jeong and Raja Jayaraman.
\newblock {Support vector-based algorithms with weighted dynamic time warping
  kernel function for time series classification}.
\newblock {\em Knowledge-Based Systems}, 75(June):184--191, 2015.

\bibitem{Zhang2010a}
Dongyu Zhang, Wangmeng Zuo, David Zhang, and Hongzhi Zhang.
\newblock {Time series classification using support vector machine with
  Gaussian elastic metric kernel}.
\newblock {\em Proceedings - International Conference on Pattern Recognition},
  pages 29--32, 2010.

\bibitem{Lods}
Arnaud Lods, Simon Malinowski, Romain Tavenard, and Laurent Amsaleg.
\newblock {Learning DTW-Preserving Shapelets}.
\newblock In {\em International Symposium on Intelligent Data Analysis}, pages
  198--209. Springer, Cham, 2017.

\bibitem{Lines2012}
Jason Lines, Luke~M. Davis, Jon Hills, and Anthony Bagnall.
\newblock {A shapelet transform for time series classification}.
\newblock In {\em Proceedings of the 18th ACM SIGKDD International Conference
  on Knowledge Discovery and Data Mining}, page 289, 2012.

\bibitem{Cover1967}
T.~Cover and P.~Hart.
\newblock {Nearest neighbor pattern classification}.
\newblock {\em IEEE Transactions on Information Theory}, 13(1):21--27, 1967.

\bibitem{Kaya2015}
H{\"{u}}seyin Kaya and Şule
  G{\"{u}}nd{\"{u}}z-{\"{O}}ʇ{\"{u}}d{\"{u}}c{\"{u}}.
\newblock {A distance based time series classification framework}.
\newblock {\em Information Systems}, 51:27--42, 2015.

\bibitem{Pekalska}
El{\.{z}}bieta P{\c{e}}kalska and Robert~P.W. Duin.
\newblock {\em {The Dissimilarity Representation for Pattern Recognition:
  Foundations and Applications}}.
\newblock 2005.

\bibitem{Chen2009}
Yihua Chen, Eric Garcia, and Maya Gupta.
\newblock {Similarity-based classification: Concepts and algorithms}.
\newblock {\em Journal of Machine Learning Research}, 10(206):747--776, 2009.

\bibitem{Paclik2001}
El{\.{z}}bieta P{\c{e}}kalska, Pavel Pacl{\'{i}}k, and Robert~P.W. Duin.
\newblock {A Generalized Kernel Approach to Dissimilarity-based
  Classification}.
\newblock {\em Journal of Machine Learning Research}, 2:175--211, 2001.

\bibitem{Graepel1999}
Thore Graepel, Ralf Herbrich, Peter Bollmann-Sdorra, and Klaus Obermayer.
\newblock {Classification on Pairwise Proximity Data}.
\newblock {\em Advances in Neural Information Processing Systems}, 11:438--444,
  1999.

\bibitem{Sakoe1978}
Hiroaki Sakoe and Seibi Chiba.
\newblock {Dynamic Programming Algorithm Optimization for Spoken Word
  Recognition}.
\newblock {\em IEEE Transactions on Acoustics, Speech and Signal Processing},
  26(1):43--49, 1978.

\bibitem{Giusti2017}
Rafael Giusti, Diego~F. Silva, and Gustavo~E.A.P.A. Batista.
\newblock {Improved time series classification with representation diversity
  and SVM}.
\newblock In {\em International Conference on Machine Learning and
  Applications}, number~1, pages 1--6, 2016.

\bibitem{Mori2017}
Usue Mori, Alexander Mendiburu, Eamonn Keogh, and Jose~A. Lozano.
\newblock {Reliable early classification of time series based on discriminating
  the classes over time}.
\newblock {\em Data Mining and Knowledge Discovery}, 31(1):233--263, 2017.

\bibitem{Rasmussen2006}
Carl Rasmussen and Cristopher Williams.
\newblock {\em {Gaussian Processes for Machine Learning}}.
\newblock 2006.

\bibitem{Chen2015a}
Yanping Chen, Eamonn Keogh, Bing Hu, Nurjahan Begum, Anthony Bagnall, Abdullah
  Mueen, and Gustavo~E.A.P.A. Batista.
\newblock {The UCR Time Series Classification Archive}.
\newblock 2015.

\bibitem{Goebel2016}
Thapanan Janyalikit, Phongsakorn Sathianwiriyakhun, Haemwaan Sivaraks, and
  Chotirat~Ann Ratanamahatana.
\newblock {An Enhanced Support Vector Machine for Faster Time Series
  Classification}.
\newblock In {\em Asian Conference on Intelligent Information and Database
  Systems}, pages 616--625, 2016.

\bibitem{Keogh2005}
Eamonn Keogh and Chotirat~Ann Ratanamahatana.
\newblock {Exact indexing of dynamic time warping}.
\newblock {\em Knowledge and information systems}, (February 2003):358--386,
  2005.

\bibitem{Goebel2015}
Brijnesh Jain and Stephan Spiegel.
\newblock {Dimension Reduction in Dissimilarity Spaces for Time Series
  Classification}.
\newblock In {\em International Workshop on Advanced Analytics and Learning on
  Temporal Data}, pages 31--46, 2015.

\bibitem{Freund1997}
Yoav Freund and Robert~E Schapire.
\newblock {A Decision-Theoretic Generalization of On-Line Learning and an
  Application to Boosting}.
\newblock {\em Computer and System Sciences}, 139:119--139, 1997.

\bibitem{Ye2009}
Lexiang Ye and Eamonn Keogh.
\newblock {Time series shapelets: A New Primitive for Data Mining}.
\newblock {\em Proceedings of the 15th ACM SIGKDD International conference on
  Knowledge Discovery and Data Mining}, page 947, 2009.

\bibitem{He2012}
Qing He, Dong Zhi, Fuzhen Zhuang, Tianfeng Shang, and Zhongzhi Shi.
\newblock {Fast time series classification based on infrequent shapelets}.
\newblock {\em Proceedings of the 11th ICMLA International Conference on
  Machine Learning and Applications}, 1:215--219, 2012.

\bibitem{Mueen2011}
Abdullah Mueen, Eamonn Keogh, and Neal Young.
\newblock {Logical-shapelets: an expressive primitive for time series
  classification}.
\newblock {\em Proceedings of the 17th ACM SIGKDD International Conference on
  Knowledge Discovery and Data Mining}, pages 1154--1162, 2011.

\bibitem{Rakthanmanon2013}
Thanawin Rakthanmanon and Eamonn Keogh.
\newblock {Fast Shapelets: A Scalable Algorithm for Discovering Time Series
  Shapelets}.
\newblock {\em Proceedings of the 13th ICDM International Conference on Data
  Mining}, pages 668--676, 2013.

\bibitem{Ye2011}
Lexiang Ye and Eamonn Keogh.
\newblock {Time series shapelets: A novel technique that allows accurate,
  interpretable and fast classification}.
\newblock {\em Data Mining and Knowledge Discovery}, 22(1-2):149--182, 2011.

\bibitem{Grabocka2014a}
Josif Grabocka, Nicolas Schilling, Martin Wistuba, and Lars Schmidt-Thieme.
\newblock {Learning time-series shapelets}.
\newblock In {\em Proceedings of the 20th ACM SIGKDD International Conference
  on Knowledge Discovery and Data Mining}, pages 392--401, 2014.

\bibitem{Bostrom2014}
Aaron Bostrom and Anthony Bagnall.
\newblock {Binary Shapelet Transform for Multiclass Time Series
  Classification}.
\newblock {\em Transactions on Large Scale Data and Knowledge Centered
  Systems}, 8800:24--46, 2014.

\bibitem{Bostrom2016}
Aaron Bostrom, Anthony Bagnall, and Jason Lines.
\newblock {Evaluating Improvements to the Shapelet Transform}.
\newblock In {\em www-bcf.usc.edu}, 2016.

\bibitem{Borg1997}
Ingwer Borg and Patrick Groenen.
\newblock {Modern Multidimensional Scaling: Theory and Applications}.
\newblock {\em Springer-Verlag}, 1997.

\bibitem{Wilson2014}
Richard~C. Wilson, Edwin~R. Hancock, El{\.{z}}bieta P{\c{e}}kalska, and
  Robert~P.W. Duin.
\newblock {Spherical and hyperbolic embeddings of data}.
\newblock {\em IEEE Transactions on Pattern Analysis and Machine Intelligence},
  36(11):2255--2269, 2014.

\bibitem{Jacobs2000}
David~W Jacobs, Daphna Weinshall, and Yoram Gdalyahu.
\newblock {Classification with Nonmetric Distances: Image Retrieval and Class
  Representation}.
\newblock {\em IEEE Transactions on Pattern Analysis and Machine Intelligence},
  22(6):583--600, 2000.

\bibitem{Hayashi2005}
Akira Hayashi, Yuko Mizuhara, and Nobuo Suematsu.
\newblock {Embedding time series data for classification}.
\newblock {\em International Workshop on Machine Learning and Data Mining in
  Pattern Recognition}, pages 356----365, 2005.

\bibitem{Mizuhara2006}
Yuko Mizuhara, Akira Hayashi, and Nobuo Suematsu.
\newblock {Embedding of time series data by using Dynamic Time Warping
  distances}.
\newblock {\em Systems and Computers in Japan}, 37(3):1--9, 2006.

\bibitem{Belkin2002}
Mikhail Belkin and Partha Niyogi.
\newblock {Laplacian Eigenmaps and Spectral Techniques for Embedding and
  Clustering}.
\newblock {\em Advances in Neural Information Processing Systems}, 14:585--591,
  2002.

\bibitem{Lichman2013}
M.~Lichman.
\newblock {UCI Machine Learning Repository}, 2013.

\bibitem{Lei2017}
Qi~Lei, Jinfeng Yi, Roman Vaculin, Lingfei Wu, and Inderjit~S. Dhillon.
\newblock {Similarity Preserving Representation Learning for Time Series
  Analysis}.
\newblock {\em arXiv:1702.03584 [cs]}, 2017.

\bibitem{Adams2004}
Colins~C. Adams.
\newblock {\em {The Knot Book: An Elementary Introduction to the Mathematical
  Theory of Knots}}.
\newblock 2004.

\bibitem{Sun2016}
Ruoyu Sun and Zhi~Quan Luo.
\newblock {Guaranteed Matrix Completion via Non-Convex Factorization}.
\newblock {\em IEEE Transactions on Information Theory}, 62(11):6535--6579,
  2016.

\bibitem{Cortes1995}
Corinna Cortes and Vladimir Vapnik.
\newblock {Support-Vector Networks}.
\newblock {\em Machine Learning}, 297:273--297, 1995.

\bibitem{Shawe-Taylor2004}
John Shawe-Taylor and Nello Cristianini.
\newblock {\em {Kernel methods for pattern analysis}}.
\newblock 2004.

\bibitem{Vapnik1998}
Vladimir Vapnik.
\newblock {\em {Statistical Learning Theory}}, volume~2.
\newblock New York, 1998.

\bibitem{Scholkopf2002}
Bernhard Sch{\"{o}}lkopf.
\newblock {\em {Learning with kernels: support vector machines, regularization,
  optimization, and beyond}}.
\newblock 2001.

\bibitem{Leslie2002}
Christina Leslie, Eleazar Eskin, and William~Stafford Noble.
\newblock {the Spectrum Kernel: a String Kernel for Svm Protein
  Classification}.
\newblock In {\em Proceedings of the Pacific Symposium on Biocomputing}, pages
  564--575, 2002.

\bibitem{Ruping2001}
Stefan R{\"{u}}ping.
\newblock {SVM Kernels for Time Series Analysis}.
\newblock Technical report, 2001.

\bibitem{Ong2004}
Cheng~Soon Ong, Xavier Mary, St{\'{e}}phane Canu, and Alexander~J. Smola.
\newblock {Learning with non-positive kernels}.
\newblock {\em Proceedings of the 21th ICML International Conference on Machine
  Learning}, (7):81, 2004.

\bibitem{Bahlmann2002}
Claus Bahlmann, Bernard Haasdonk, and Hans Burkhardt.
\newblock {Online handwriting recognition with support vector machines - A
  kernel approach}.
\newblock {\em Proceedings - International Workshop on Frontiers in Handwriting
  Recognition, IWFHR}, pages 49--54, 2002.

\bibitem{Decoste2002}
Dennis Decoste and Bernhard Sch{\"{o}}lkopf.
\newblock {Training Invariant Support Vector Machines using Selective
  Sampling}.
\newblock {\em Machine Learning}, 46,:161--190, 2002.

\bibitem{Shimodaira2006}
Hiroshi Shimodaira, Ken~Ichi Noma, Mitsuru Nakai, and Shigeki Sagayama.
\newblock {Dynamic Time-Alignment Kernel in Support Vector Machine.}
\newblock {\em Advances in Neural Information Processing Systems},
  2(1):921--928, 2002.

\bibitem{Haasdonk2005}
Bernard Haasdonk.
\newblock {Feature space interpretation of SVMs with indefinite kernels}.
\newblock {\em IEEE Transactions on Pattern Analysis and Machine Intelligence},
  27(4):482--492, 2005.

\bibitem{Chen2006}
Pai-hsuen Chen, Rong-en Fan, and Chih-jen Lin.
\newblock {A Study on SMO-Type Decomposition Methods for Support Vector
  Machines}.
\newblock {\em IEEE Transactions on Neural Networks and Learning Systems},
  17(4):893--908, 2006.

\bibitem{Haasdonk2004}
Bernard Haasdonk and Claus Bahlmann.
\newblock {Learning with Distance Substitution Kernels}.
\newblock {\em Pattern Recognition}, pages 220--227, 2004.

\bibitem{Pree2014a}
Helmuth Pree, Benjamin Herwig, Thiemo Gruber, Bernhard Sick, Klaus David, and
  Paul Lukowicz.
\newblock {On general purpose time series similarity measures and their use as
  kernel functions in support vector machines}.
\newblock {\em Information Sciences}, 281:478--495, 2014.

\bibitem{Jeong2011}
Young-seon Jeong, Myong~K Jeong, and Olufemi~A Omitaomu.
\newblock {Weighted dynamic time warping for time series classification}.
\newblock {\em Pattern Recognition}, 44(9):2231--2240, 2011.

\bibitem{Chen2015}
Zhihua Chen, Wangmeng Zuo, Qinghua Hu, and Liang Lin.
\newblock {Kernel sparse representation for time series classification}.
\newblock {\em Information Sciences}, 292:15--26, 2015.

\bibitem{Lei2007}
Hansheng Lei and Bingyu Sun.
\newblock {A Study on the Dynamic Time Warping in Kernel Machines}.
\newblock {\em Proceedings of the 3rd SITIS International IEEE Conference on
  Signal-Image Technologies and Internet-Based System}, pages 839--845, 2007.

\bibitem{Kaya2013}
H{\"{u}}seyin Kaya and Şule
  G{\"{u}}nd{\"{u}}z-{\"{O}}ʇ{\"{u}}d{\"{u}}c{\"{u}}.
\newblock {SAGA: A novel signal alignment method based on genetic algorithm}.
\newblock {\em Information Sciences}, 228:113--130, 2013.

\bibitem{Cuturi2011}
Marco Cuturi.
\newblock {Fast Global Alignment Kernels}.
\newblock In {\em Proceedings of the 28th ICML International Conference on
  Machine Learning}, pages 929--936, 2011.

\bibitem{Guyon1994}
Isabelle Guyon, Lambert Schomaker, Rkjean Planiondon, Mark Liberman, Stan
  Janet, Ecole Polytechnique~De Montreal, and Linguistic~Data Consortium.
\newblock {UNIPEN project of on-line data exchange}.
\newblock pages 29--33, 1994.

\bibitem{Hochreiter2006}
Sepp Hochreiter and Klaus Obermayer.
\newblock {Support Vector Machines for Dyadic Data}.
\newblock {\em Neural Computation}, 1510:1472--1510, 2006.

\bibitem{Wu2005}
Gang Wu, Edward~Y. Chang, and Zhihua Zhang.
\newblock {An analysis of transformation on non-positive semidefinite
  similarity matrix for kernel machines}.
\newblock {\em Proceedings of the 22th ICML International Conference on Machine
  Learning}, 8, 2005.

\bibitem{Wu2005a}
Gang Wu, Edward~Y. Chang, and Zhihua Zhang.
\newblock {Learning with non-metric proximity matrices}.
\newblock {\em Proceedings of the 13th ACM International Conference on
  Multimedia}, page 411, 2005.

\bibitem{Zhang2012}
Li~Zhang, Pei-chann Chang, Jing Liu, Zhe Yan, Ting Wang, and Fan-zhang Li.
\newblock {Kernel Sparse Representation-Based Classifier}.
\newblock {\em IEEE Transactions on Signal Processing}, 60(4):1684--1695, 2012.

\bibitem{Chen2004}
Lei Chen and Raymond Ng.
\newblock {On The Marriage of Lp-norms and Edit Distance}.
\newblock In {\em International conference on Very large data bases}, pages
  792--803, 2004.

\bibitem{Marteau2009}
Pierre-Fran{\c{c}}ois Marteau.
\newblock {Time Warp Edit Distance with Stiffness Adjustment for Time Series
  Matching}.
\newblock {\em IEEE Transactions on Pattern Analysis and Machine Intelligence},
  31(2):306--318, 2009.

\bibitem{Weston2003}
Jason Weston, Bernhard Sch{\"{o}}lkopf, Eleazar Eskin, Christina Leslie, and
  William~Stafford Noble.
\newblock {Dealing with large diagonals in kernel matrices}.
\newblock In {\em Annals of the Institute of Statistical Mathematics},
  volume~55, pages 391--408, 2003.

\bibitem{Casacuberta1987}
F~Casacuberta, E~Vidal, and H~Rulot.
\newblock {On the metric properties of dynamic time warping}.
\newblock {\em IEEE Transactions on Acoustics, Speech and Signal Processing},
  35(11):1631--1633, 1987.

\bibitem{Tax2004}
David~M.J. Tax and Robert~P.W. Duin.
\newblock {Support Vector Data Description}.
\newblock {\em Machine Learning}, 54,:45--66, 2004.

\bibitem{Gaidon2011}
Adrien Gaidon, Zaid Harchoui, and Cordelia Schmid.
\newblock {A time series kernel for action recognition}.
\newblock In {\em Procedings of the British Machine Vision Conference}, pages
  63.1--63.11, 2011.

\bibitem{Troncoso2015}
A.~Troncoso, M.~Arias, and J.~C. Riquelme.
\newblock {A multi-scale smoothing kernel for measuring time-series
  similarity}.
\newblock {\em Neurocomputing}, 167:8--17, 2015.

\bibitem{Kumara2008}
Karthik Kumara, Rahul Agrawal, and Chiranjib Bhattacharyya.
\newblock {A large margin approach for writer independent online handwriting
  classification}.
\newblock {\em Pattern Recognition Letters}, 29(7):933--937, 2008.

\bibitem{Sivaramakrishnan2004}
K~R Sivaramakrishnan and Chiranjib Bhattacharyya.
\newblock {Time Series Classification for Online Tamil Handwritten Character
  Recognition – A Kernel Based Approach}.
\newblock In {\em International Conference on Neural Information Processing},
  pages 800--805, 2004.

\bibitem{Lu2008}
Zhengdong Lu, K.~Todd Leen, Yonghong Huang, and Deniz Erdogmus.
\newblock {A Reproducing Kernel Hilbert Space Framework for Pairwise Time
  Series Distances}.
\newblock In {\em Proceedings of the 25th ICML International Conference on
  Machine learning}, volume~56, pages 624--631, 2008.

\bibitem{Xue2017}
Yangtao Xue, Li~Zhang, Zhiwei Tao, Bangjun Wang, and Fan-zhang Li.
\newblock {An Altered Kernel Transformation for Time Series Classification}.
\newblock In {\em International Conference on Neural Information Processing},
  pages 455--465, 2017.

\bibitem{Wachman2009}
Gabriel Wachman, Roni Khardon, Pavlos Protopapas, and Charles {R. Alcock}.
\newblock {Kernels for Periodic Time Series Arising in Astronomy}.
\newblock In {\em European Conference on Machine Learning and Knowledge
  Discovery in Databases}, 2009.

\bibitem{Marteau2010}
Pierre-Fran{\c{c}}ois Marteau and Sylvie Gibet.
\newblock {Constructing Positive Definite Elastic Kernels with Application to
  Time Series Classification}.
\newblock {\em CoRR}, pages 1--18, 2010.

\bibitem{Marteau2012}
Pierre-Fran{\c{c}}ois Marteau, Nicolas Bonnel, and Gilbas M{\'{e}}nier.
\newblock {Discrete Elastic Inner Vector Spaces with Application in Time Series
  and Sequence Mining}.
\newblock {\em IEEE Transactions on Knowledge and Data Engineering},
  25(9):2024--2035, 2012.

\bibitem{Greub1975}
W.~H. Greub.
\newblock {\em {Linear algebra}}.
\newblock Springer-Verlag, 1975.

\bibitem{Cortes2004}
Corinna Cortes, Patrick Haffner, and Mehryar Mohri.
\newblock {Rational Kernels: Theory and Algorithms}.
\newblock {\em Journal of Machine Learning Research}, 5:1035--1062, 2004.

\bibitem{Wu2018}
Lingfei Wu, Ian En-Hsu Yen, Fangli Xu, Pradeep Ravikuma, and Michael Witbrock.
\newblock {D2KE: From Distance to Kernel and Embedding}.
\newblock {\em arxiv.org/abs/1802.04956v3}, pages 1--18, 2018.

\bibitem{Liberman1993}
Mark Liberman.
\newblock {TI46 speech corpus}.
\newblock In {\em Linguistic Data Consortium}, 1993.

\bibitem{Serra2014}
Joan Serr{\`{a}} and Josep~Ll Arcos.
\newblock {An empirical evaluation of similarity measures for time series
  classification}.
\newblock {\em Knowledge-Based Systems}, 67:305--314, 2014.

\bibitem{Pekalska2006}
El{\.{z}}bieta P{\c{e}}kalska, Robert~P.W. Duin, and Pavel Pacl{\'{i}}k.
\newblock {Prototype selection for dissimilarity-based classifiers}.
\newblock {\em Pattern Recognition}, 39(2):189--208, 2006.

\end{thebibliography}
\end{document}